\documentclass{article}
\makeatletter
\renewcommand*{\@fnsymbol}[1]{\ensuremath{\ifcase#1\or *\or **\or \dagger\or \ddagger\or
   \mathsection\or \mathparagraph\or \|\or \dagger\dagger
   \or \ddagger\ddagger \else\@ctrerr\fi}}
\makeatother

\usepackage{geometry}
\usepackage{amsmath,amssymb,amsthm}
\usepackage{todonotes}

\usepackage{algorithm}
\usepackage{algpseudocode}

\usepackage{graphicx}
\graphicspath{{images/}}

\usepackage{subcaption}
\usepackage[labelfont=sc]{caption}

\usepackage[mode=buildnew]{standalone}

\usepackage{hyperref}

\usepackage{csvsimple}

\newcommand{\Scal}{\mathcal{S}}
\newcommand{\Mcal}{\mathcal{M}}
\newcommand{\Acal}{\mathcal{A}}
\newcommand{\Ucal}{\mathcal{U}}

\newcommand{\eps}{\varepsilon}
\newcommand{\sopt}{s^\star}

\newcommand{\nls}{\mathcal{S}_\mathrm{next}}

\numberwithin{equation}{section}

\begin{document}

\title{Knowledge State Networks for Effective Skill Assessment in Atomic Learning}
\author{Julian Rasch\thanks{edyoucated GmbH (julian@edyoucated.org, david@edyoucated.org)}, 
  David Middelbeck$^*$}
\maketitle
\begin{abstract}
The goal of this paper is to introduce a new framework for fast and effective knowledge state assessments in the context of personalized, skill-based online learning.  
We use knowledge state networks -- specific neural networks trained on assessment data of previous learners -- to predict the full knowledge state of other learners from only partial information about their skills.
In combination with a matching assessment strategy for asking discriminative questions we demonstrate that our approach leads to a significant speed-up of the assessment process -- in terms of the necessary number of assessment questions -- in comparison to standard assessment designs.
In practice, the presented methods enable personalized, skill-based online learning also for skill ontologies of very fine granularity without deteriorating the associated learning experience by a lengthy assessment process.
\end{abstract}


\section{Introduction}
The need for effective re- and upskilling opportunities has never been larger than today. 
More than one billion jobs worldwide are expected by the \cite{WEF} to be fundamentally transformed by technology in this decade, leading to an unprecedented need for new training and lifelong learning opportunities \cite{Field2000,Knapper2000,Jarvis2004,Jarvis2014,Yang2015,Kaplan2017,Beblavy2019,Alt2020}.
To address this strong demand for new and scalable learning opportunities, there has been a significant shift towards online learning in the past years – a development that has been further fueled by the COVID-19 pandemic. 

This has led to a substantial increase in online learning resources over the last decade, and in particular drove the development of online platforms distributing massive open online courses (MOOCs) \cite{Daradoumis2013,Fournier2015,Kiselev2020}. 
While MOOCs provide autonomous, flexible and independent education at high quality and often relatively low cost \cite{Daradoumis2013,Saadatdoost2015,StClair2015}, their main focus lies on a high scalability; the goal is to provide learning content for the broadest possible range of learners.
This, however, often leads to trainings and courses that follow one-fits-all strategies \cite{McBride2004,Murray2004,Gasevic2016} and do not take into account the different qualifications, needs and learning goals of their participants.
And indeed, the measurable success of MOOCs has been rather limited so far; depending on the source, completion rates have been reported to often lie between five and fifteen percent only \cite{Perna2013,Banerjee2014,Guetl2014,Khalil2014}.
The reasons for drop-out are diverse, ranging from insufficient time allocation, difficulties with the subject matter or unchallenging activities to disorganization and insufficient planning \cite[for example]{Guetl2014,Khalil2014}. 
While there has been a substantial amount of research on the use of machine learning for the automated prediction of drop-out from features predominantly revolving around learner activity \cite[for example]{Fei2015,Xing2016,Dalipi2018}, there has only been very recent work on implementing actual interventions to prevent drop-out and investigating their effect \cite{Borrella2019,Xing2019,Goopio2020}. 

A similar observation can be made when looking at corporate learning and development initiatives.
It is long known that companies need to develop into so-called learning organizations in order to face the challenge of fast restructuring due to global competition, new knowledge and technology \cite{Marquardt2011,Ben2015}.
Driven by the economical and technological advantages as well as flexibility we experience a similar shift from traditional face-to-face learning to various forms of e-learning and blended learning \cite{Macpherson2004,Newton2007,Little2016}.
While the studies of \cite{Masie2012} show a stronger similarity between higher education and corporate learning there is only little evidence that online learning or the implementation of MOOCs show better completion rates or success in a corporate setting \cite[see for some rare evaluations]{Rodriguez2013,Malca2015,Beinicke2018}. 

In the light of the rising need for lifelong learning, a primary challenge for businesses and educational institutions is the growing heterogeneity of their learner base \cite{Pitts2010,Alcazar2013}. 
The modern workplace is characterized by up to five generations of people working together, each bringing a vastly different set of talent, skills, beliefs and expectations \cite{OECD2010,Meister2021}.
New generations of employees enter their working life with an entirely different perception and use of technology than their older colleagues and employers \cite{Meister2021}. 
Organizations increasingly hire new talent from all around the world, bringing new cultures, working styles, knowledge, expectations and skills to the workplace \cite{Marquardt2011,Meister2021}.
As a consequence, compared to learners in high schools or universities, the variety of backgrounds, experiences, prior knowledge, and learning goals is significantly larger when it comes to up- and reskilling in a business environment. 
And indeed, \cite{Khalil2014} identify the issue of insufficient prior knowledge and background skills as a particularly interesting and common cause for drop-out in MOOCs.
Simply speaking, learners either lack the necessary prior knowledge to get started comfortably such that they are overwhelmed by the subject matter or, vice versa, lose interest when unnecessarily repeating content they already know.

Consequently, \cite{ODonnell2015} state that the personalization of online courses has become a field of increasing interest and research for the educational community as well as for businesses and employers. 
The key to dealing with this diversity of learners is the \textit{personalization of learning}: adaptive methods that provide appropriate and effective learning strategies to the different learners on an individual level \cite{Nelson2007,Zajac2009,Paquette2015,Yu2017,Assami2018,Yau2020}.
Personalization can be carried out in a multitude of ways, targeting the personal learning preferences \cite{Loo2004,Murphy2004,Smith2005}, cultural traits and personality \cite{Rai2016,Makhija2018,Ruedian2019} such as the ``Big Five'' personality traits \cite{McCrae1985,Goldberg1992,Komarraju2011} or the learning orders and the related content itself \cite{Garrido2014,Raghuveer2014,Assami2018,Joy2019}.
These personalization methods typically rely on analytical data, sometimes called digital traces or log data (e.g., \cite{Gasevic2015}), which is gathered for learning analytics \cite[for example]{Elias2011,Dawson2014,Larrabee2019} before, during and after the learning process in order to both understand and optimize learning itself and the related learning environments \cite{Gasevic2017}.

In this paper we focus on the skill-based personalization of the learning content depending on the prior knowledge of learners. With accurate information about the current knowledge of the learner, we can then start the learning process individually at the right place (in the sense of knowledge) for each learner, recommending the individual best next skills to learn in order to master the chosen learning topic, thereby engaging and challenging the learner in an appropriate fashion.
This type of personalization addresses the above mentioned problems of online learning in several ways:
Learners with little to no prior knowledge can start with the basics at the very beginning of the learning topic, while more experienced learners are able to skip some part of the content.
In case a learner already has prior knowledge in the subject, the saved time can be allocated to different (learning) tasks, thus also effectively tackling the issue of time as a bottleneck for MOOCs discovered by \cite{Eriksson2017}.
In this context it is also worth mentioning the concept of mastery-based learning (or mastery learning) which follows a related instructional approach by only allowing students to progress to subsequent topics once they demonstrated sufficient knowledge in the current topic \cite{Block1976,Jazayeri2015,McGaghie2015,West2015}.

Most previous work (see the references above) as well as most popular learning platforms address skills-based "personalization" by giving recommendations for \textit{entire courses} (or parts thereof) to be learned. 
Thus, personalization often happens on an insufficiently low level of granularity (e.g., recommending a 40-hour "Microsoft Excel Intermediate" course for a learner with some initial knowledge in Excel) – resulting in low effectiveness and high dropout rates. 

In contrast to previous work, the personalization framework \textsc{POLARIS}\footnote{an acronym for ``Personalized Online Learning via Atomic Recommendation and Inference of Skills''} introduced by the authors of this paper allows learning platforms to recommend single skills: small, learnable units of knowledge to be learned next, resulting in a highly individualized learning path (i.e., personalization on a 5-to-15-minute-level instead of a multi-hour-level).
To be as precise as possible with our recommendation, we use \textit{atomic skills}, i.e., we break down the subject matter into the most elementary bricks possible, offering the possibility to recommend and learn skills on a very fine granularity. 
These atomic skills are stored in skill ontologies \cite[for example]{Reich2002,Paquette2007,Askar2009} in order to be accessed and used by algorithms, which often comprise several hundred individual items.
To teach the individual skills, learning materials are attached to each skill and can be distributed to the learners.
The methods we develop and their terminology are partly inspired by the idea of knowledge and learning spaces originally introduced by Doignon and Falmagne \cite{Falmagne2006,Falmagne2011,Doignon2012}, however, take a different approach at their definition and use.

Quite obviously, the task of giving a good recommendation for the next skill requires to gather information about the current knowledge state (i.e., the complete set of skills a learner has mastered or not) in form of an assessment. 
However, on such a fine granularity with possibly multiple hundreds of skills, such an assessment can turn into a tedious task. 
In case of a self-assessment, the learner is forced to answer a large amount of questions (worst case: one question per atomic skill) before being able to start with the learning process. 
In most practical learning situations, this is almost impossible or at least a major limitation.

To this end, we propose the use of a \textit{knowledge state network}, a neural network which is able to predict the most likely knowledge states of learners throughout the assessment process, based on the assessments of previous learners of the topic. 
During our validation experiments, the  network has been implemented and tested with real learner data and with different learning topics on the personalized learning platform \textsc{edyoucated.org}. 
Combined with an intelligent assessment strategy (i.e., the order in which single atomic skills are assessed) we show that the use of such networks can reliably and significantly speed up the assessment process, making atomic skill recommendations viable and effective in practice.
Simply speaking, the proposed approach enables us to determine the full knowledge state of a learner by asking relatively few questions, such that the optimal next skill to be learned can be recommended.

\section{Atomic ontologies, knowledge states and assessment}
In the following we line out the concept of ontologies and clarify the idea behind \textit{atomic} ontologies, before we use the introduced concept to define what we understand by knowledge states and assessment states. 

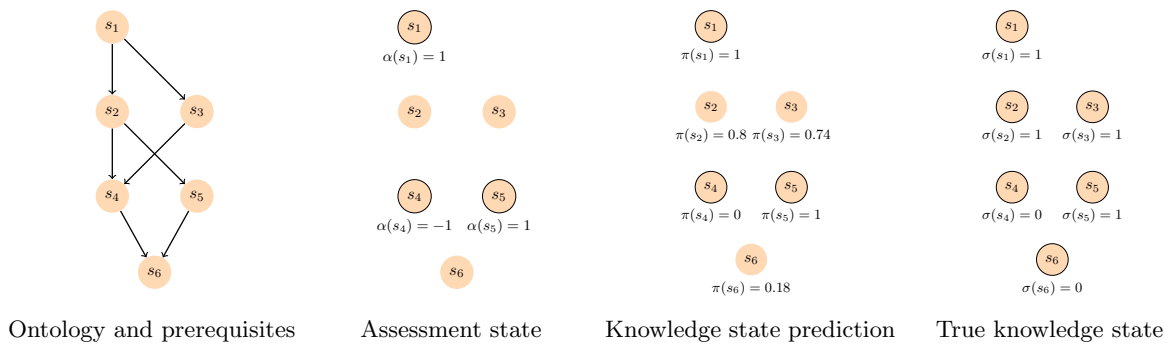
\begin{figure*}[!t]
    \centering
    \captionsetup[subfigure]{labelformat=empty, justification=centering}
    \begin{subfigure}[c]{0.24\textwidth}
        \centering
        \newcommand{\dist}{1.8}

\begin{tikzpicture}

\node[circle, minimum size = 6mm, fill=orange!30] (A) at (0, 0) {$s_1$};
\node[circle, minimum size = 6mm, fill=orange!30] (B) at (0, -\dist) {$s_2$};
\node[circle, minimum size = 6mm, fill=orange!30] (C) at (\dist, -\dist) {$s_3$};
\node[circle, minimum size = 6mm, fill=orange!30] (D) at (0, -2*\dist) {$s_4$};
\node[circle, minimum size = 6mm, fill=orange!30] (E) at (\dist, -2*\dist) {$s_5$};
\node[circle, minimum size = 6mm, fill=orange!30, label={-90:$ $}] (F) at (\dist/2, -2.9*\dist) {$s_6$};

\path[->,draw,thick] 
(A) edge (B)
(A) edge (C)
(B) edge (D)
(B) edge (E)
(C) edge (D)
(D) edge (F)
(E) edge (F)
;

\end{tikzpicture}
        \subcaption{Ontology and prerequisites}
    \end{subfigure}     
    \begin{subfigure}[c]{0.24\textwidth}
        \centering
        \newcommand{\dist}{1.8}

\begin{tikzpicture}

\node[circle, minimum size = 6mm, fill=orange!30, draw=black, label={-90:\small$\alpha(s_1)=1$}] (A) at (0, 0) {$s_1$};
\node[circle, minimum size = 6mm, fill=orange!30] (B) at (0, -\dist) {$s_2$};
\node[circle, minimum size = 6mm, fill=orange!30] (C) at (\dist, -\dist) {$s_3$};
\node[circle, minimum size = 6mm, fill=orange!30, draw=black, label={-90:\small$\alpha(s_4)=-1$}] (D) at (0, -2*\dist) {$s_4$};
\node[circle, minimum size = 6mm, fill=orange!30, draw=black, label={-90:\small$\alpha(s_5)=1$}] (E) at (\dist, -2*\dist) {$s_5$};
\node[circle, minimum size = 6mm, fill=orange!30, label={-90:$ $}] (F) at (\dist/2, -2.9*\dist) {$s_6$};

\end{tikzpicture}
        \subcaption{Assessment state}
    \end{subfigure}     
    \begin{subfigure}[c]{0.24\textwidth}
        \centering
        \newcommand{\dist}{1.8}

\begin{tikzpicture}

\node[circle, minimum size = 6mm, fill=orange!30, draw=black, label={-90:\small$\pi(s_1)=1$}] (A) at (0, 0) {$s_1$};
\node[circle, minimum size = 6mm, fill=orange!30, label={-90:\small$\pi(s_2)=0.8$}] (B) at (0, -\dist) {$s_2$};
\node[circle, minimum size = 6mm, fill=orange!30, label={-90:\small$\pi(s_3)=0.74$}] (C) at (\dist, -\dist) {$s_3$};
\node[circle, minimum size = 6mm, fill=orange!30, draw=black, label={-90:\small$\pi(s_4)=0$}] (D) at (0, -2*\dist) {$s_4$};
\node[circle, minimum size = 6mm, fill=orange!30, draw=black, label={-90:\small$\pi(s_5)=1$}] (E) at (\dist, -2*\dist) {$s_5$};
\node[circle, minimum size = 6mm, fill=orange!30, label={-90:\small$\pi(s_6)=0.18$}] (F) at (\dist/2, -2.9*\dist) {$s_6$};

\end{tikzpicture}
        \subcaption{Knowledge state prediction}
    \end{subfigure} 
    \begin{subfigure}[c]{0.24\textwidth}
        \centering
        \newcommand{\dist}{1.8}

\begin{tikzpicture}

\node[circle, minimum size = 6mm, fill=orange!30, draw=black, label={-90:\small$\sigma(s_1)=1$}] (A) at (0, 0) {$s_1$};
\node[circle, minimum size = 6mm, fill=orange!30, draw=black, label={-90:\small$\sigma(s_2)=1$}] (B) at (0, -\dist) {$s_2$};
\node[circle, minimum size = 6mm, fill=orange!30, draw=black, label={-90:\small$\sigma(s_3)=1$}] (C) at (\dist, -\dist) {$s_3$};
\node[circle, minimum size = 6mm, fill=orange!30, draw=black, label={-90:\small$\sigma(s_4)=0$}] (D) at (0, -2*\dist) {$s_4$};
\node[circle, minimum size = 6mm, fill=orange!30, draw=black, label={-90:\small$\sigma(s_5)=1$}] (E) at (\dist, -2*\dist) {$s_5$};
\node[circle, minimum size = 6mm, fill=orange!30, draw=black, label={-90:\small$\sigma(s_6)=0$}] (F) at (\dist/2, -2.9*\dist) {$s_6$};

\end{tikzpicture}
        \subcaption{True knowledge state}
    \end{subfigure}  
    \caption{Assessment states, predicted knowledge states and true knowledge states for an example ontology with seven atomic skills.
    Skills with known state are circled.}
    \label{fig:ontology_states}
\end{figure*}

\subsection{Atomic ontologies}\label{subsec:atomic_ontologies}
At its core, an ontology is a means to model the structure of some (complex) system in a formal way \cite{Staab2010}.
It comprises the relevant entities of the system, which are connected by relations of all kinds in order to form dependencies between the different entities, or groups and hierarchies among them. 
The easiest way to visualize an ontology (at least, in our setting), is to think of a directed graph consisting of a number of nodes representing the entities of the system connected by edges modeling the relationship between nodes (cf. Figure \ref{fig:ontology_states}).

In the field of education, the ontology of a topic comprises all the individual skills that can be acquired by a learner, from the easiest to the most complicated subject, which form (some of) the nodes of the graph (see, e.g., \cite{Falmagne2006} for a nice representation). 
The relations between skills are used to model useful connections, such as an appropriate order in which the skills should be learned, prerequisite relations (see below) or the grouping of skills into coarser concepts.
Depending on the use case and the necessary or desired complexity, the granularity of ontologies can be rather coarse or fine.
While, for example, \cite{Ibrahim2019} found that for ontology-based course recommendation the structure can be rather coarse, it is required to be of much finer granularity to really capture the different elements of a learning curriculum \cite[p. 7]{Falmagne2006}.

At edyoucated\footnote{\url{www.edyoucated.org}}, we use what we call \textit{atomic ontologies}, where the idea is to break down a particular topic into its elementary bricks, i.e., into the smallest units of knowledge that can be learned individually (which, of course, depend on each other). 
Two simple examples for such atomic skills are the \textit{for-loop in Python} or the \textit{summation formula in MS Excel}. 
It is clear that these skills cannot be broken down any further; they have prerequisites, e.g., the definition of variables for a for loop, but the for-loop itself cannot reasonably be split further into any smaller ``sub-topics''.

With the ontology at hand, the goal of a skill assessment (e.g., \cite{Weeden2002}) is now to figure out, which parts of the ontology are already known to a learner and which parts are not.
The assessment can take different forms, where the most common are either self-assessments or knowledge tests in forms of exercises.
When it comes to the assessment of skills, it is easy to find a notable difference between atomic skills and coarser skills. 
Due to their coarser definition, the latter usually require the learner to rank their expertise in a certain topic, such as \textit{Python Fundamentals} or \textit{Microsoft Excel Formulas}, on an ordinal scale, which is often not a trivial task for the learner.
In contrast, atomic learning allows for a rather simple style of assessment, where the learner can answer in a binary fashion; they have either already mastered the skill or not.
Due to the fine granularity of skills, knowledge does not have to be located on a scale, which makes the (self-)assessment comparably easier.
The edyoucated platform offers the possibility to store and access such atomic ontologies and use them for content personalization in practice, allowing the learners to acquire personalized knowledge in small, manageable bits one at a time.

In order to turn a collection of (atomic) skills $\Scal = \{s_1, \dots, s_n\}$ into an ontology, we need to define relations between the skills, describing properties that connect certain skills to others.
A rather obvious choice for learning is to connect the skills via a precedence or prerequisite relation (\cite{Falmagne2006}, cf. also Figure \ref{fig:ontology_states}). 
The key behind this is the observation that, in order to master a certain set of skills, a learner necessarily will have to have acquired knowledge about other skills: 
the skill prerequisites.
These relations usually have to be modeled manually by experts which requires a substantial amount of work for bigger sets of skills.

Prerequisites offer a convenient way to perform skill assessments: 
Assume that a learner knows skill $s_5$ in Figure \ref{fig:ontology_states}. 
The prerequisite relations then recursively imply that the skills $s_2$ and $s_1$ must have been known as well, such that we do not have to assess these skills explicitly. 
Vice versa, if $s_5$ is unknown, we immediately deduce that $s_6$ cannot be known either. 
We refer to \cite{Falmagne2006} and \cite{Doignon2012} for an algorithm which is able to pick the right order of questions for an assessment in order to exploit this structure and produce a relatively short assessment (we present alternatives in a slightly different setting later in this paper). 

Despite their convenience for assessments, prerequisite relations feature another notable disadvantage in addition to the manual modeling effort for their creation.
Oftentimes it is difficult to decide whether a prerequisite is strict, implying that consecutive skills definitely \textit{cannot} be learned without mastery of the skill at hand, or whether it is only a soft prerequisite where subsequent skills can potentially be mastered even when skipping the skill.
The modeling of prerequisites is hence often rather subjective and depends heavily on the expert that performed the task. 
Of similar difficulty is the subtle difference between a skill B requiring knowledge about a second skill A and the mastery of B \textit{actually} implying mastery of A, which theoretically requires to model multiple relations. 
All this, together with the fact that the manual modeling becomes increasingly impossible with a large and growing number of skills for atomic learning, makes the approach impracticable in the long run.

One of the main contributions of this paper is hence a method to implicitly \textit{learn} these relations from user data, more precisely, from the assessment of previous learners.

\subsection{Knowledge and assessment states}
The key ingredient for the (automatic) deduction of relations between skills is the notion of a knowledge state, originally introduced in \cite{Falmagne2006} and \cite{Doignon2012}, even though we are defining it in a different way. 
As mentioned above, a knowledge state is a representation of the knowledge of a learner, letting us know, which of the skills of our ontology have already been mastered or are still unknown.
The goal is to predict such a knowledge state from an incomplete version that we call assessment state.
In general, the task can be formulated as follows: 
Given the assessments of a few skills of a learner, what is the most probable full knowledge state?

In order to fix the terminology, let $\Scal = \{s_1, \dots, s_n\}$ be a set of $n$ skills (i.e., our ontology). 
For the sake of notational ease we model any skill-related information about a learner (a \textit{state}) as a function which changes its value with time, representing the learning or assessment progress of a learner.\footnote{By a slight abuse of notation we will leave out the time parameter $t$ for brevity whenever the considered time point is clear or irrelevant.}
The first state we define is the \textit{knowledge state} of a learner which, for each skill $s$ and time $t$ is a mapping $\sigma \colon \Scal \times [0, \infty) \to \{0, 1\}$ such that 
\begin{align*}
    \sigma(s,t) = 
    \begin{cases}
        1, & \text{learner knows $s$ at time $t$}, \\
        0, & \text{learner does not know $s$ at time $t$}.
    \end{cases}
\end{align*}
The starting time $t = 0$ of the time domain may, for example, be interpreted as the moment a learner enters the learning platform for the first time. 
In general, however, the starting point of the time domain can be adjusted to the particular application.
If the knowledge state of the learner changes between any two time points $t_k < t_{l}$, for instance, if the learner gains knowledge about a previously unknown skill $s \in \Scal$, then $\sigma(s, t_k) = 0$ and $\sigma(s, t_l) = 1$.
The knowledge state of a learner for the entire set of skills $\Scal$ at time $t$ is then given as
\begin{align*}
    \sigma(\Scal, t) = \{\sigma(s, t) \,|\, s \in \Scal \} \in \Sigma,
\end{align*}
and is an element of the set $\Sigma$ of all possible knowledge states.
Note that $\Sigma$ itself indeed does not depend on $t$, as it is simply the deterministic set of all knowledge permutations of the skills in $\Scal$.

The skill assessment related to $\Scal$ can now be thought of as an iterative process over time, where in each iteration we gain some additional information about the learner's knowledge state. 
Hence, analogously to a knowledge state, the \textit{assessment state} of a learner for a skill $s$ can be expressed as a mapping $\alpha \colon \Scal \times [0, \infty) \to \{-1, 0, 1\}$ such that 
\begin{align*}
    \alpha(s, t) = 
    \begin{cases}
        \phantom{-}1 & \text{if } \sigma(s, t) = 1, \\
        \phantom{-}0 & \text{knowledge state of $s$ unknown at $t$}, \\
        -1 & \text{if } \sigma(s, t) = 0.
    \end{cases}
\end{align*}
The assessment state of a learner for the set of skills $\Scal$ at time $t$ is then
\begin{align*}
    \alpha(\Scal, t) = \{\alpha(s, t) \,|\, s \in \Scal \} \in \Acal,
\end{align*}
and we denote the set of all assessment states by $\Acal$.
We call an assessment state \textit{incomplete}, if $\alpha(s, t) = 0$ for at least one $s \in \Scal$.
Note the subtle difference between an assessment state and a knowledge state:
The true knowledge state of a learner usually stays a fixed entity (mathematically, a constant function) throughout the assessment process, as the learner does not gain any knowledge during that time.
The assessment state, on the contrary, changes.
With every additional step of the assessment, we gain definitive information about one additional, previously unassessed skill $s$, such that $\alpha$ differs from zero for this particular $s$.

The knowledge state of a skill is purposefully modeled such that it can be interpreted as a probability, stating that a learner either fully mastered the skill or does not know anything about it.
In that sense, it might make sense to deviate from a binary knowledge state towards a more flexible knowledge state in certain settings, where $\sigma(s) \in [0, 1]$ expresses the level or amount of knowledge a learner possesses of a skill. 
This, however, leads to rather complicated self-assessments for the learner which makes it difficult to use in practice. 
Furthermore, in the situation of atomic learning, the fine granularity of a skill should almost never allow to have a certain level of knowledge in a skill but rather simply having it mastered or not (cf. Sec. \ref{subsec:atomic_ontologies}). 
It is worth noting that the cardinality of the set of all possible knowledge states $\Sigma$ and possible assessment states is given by $|\Sigma| = 2^n$ respectively $|\Acal| = 3^n$ and hence increases the complexity exponentially with the number of skills to be assessed.
We shall see later, however, that this does not impose any problems in practice.

\section{Knowledge state networks}
In this section and the following one we want to line out our approach to atomic skill assessments, using what we call a knowledge state network to predict the knowledge state of learners from incomplete assessment states. 
Together with a smart choice of assessment order (see Section \ref{sec:assessment_strategies}) this can lead to a substantial speed-up in assessment time while maintaining a high accuracy. 
This approach is part of the larger AI-based personalization framework \textsc{POLARIS}\footnote{an acronym for ``Personalized Online Learning via Atomic Recommendation and Inference of Skills''} that the authors of this paper have developed over the past years.

\subsection{Building a knowledge state prediction model}
The goal is to determine a learner's knowledge state with as little assessment effort as possible, i.e., with as few assessment questions as possible.
In order to achieve this we need to create a model which is able to predict missing parts of a learner's knowledge state using only partial information. 
Simply speaking, given the knowledge state of $k < n$ skills, the model has to predict the state of the missing $n-k$ skills.

Mathematically, a perfect prediction model $\hat{\Mcal} \colon \Acal \to \Sigma$ takes any (in particular, incomplete) assessment state and maps it to the corresponding knowledge state of a learner. 
It is clear that, in practice, such a model does not exist; we cannot expect a perfect prediction from arbitrarily incomplete assessment states.
We hence aim to build a probabilistic model $\Mcal \colon \Acal \to [0, 1]^n$ mapping any assessment state $A \in \Acal$ to a set of probabilities, for each of the skills representing the probability of the learner knowing it.
For each assessment state, the model $\Mcal$ hence induces a function
\begin{align*}
    \pi_\Mcal \colon \Scal \to [0, 1],
\end{align*}
mapping any skill to its current probability.
For the sake of brevity we will leave out the subscript $\Mcal$ whenever it is obvious how $\pi_\Mcal$ is defined.
It is clear that for such a model $\Mcal$, an assessment state and its induced probability function $\pi$ we immediately have
\begin{align*}
    \pi(s) = 
    \begin{cases}
        0, & \text{if } \alpha(s) = -1, \\
        1, & \text{if } \alpha(s) = 1,    
    \end{cases}
\end{align*}
which simply expresses our exact knowledge about the skill state of the learner when the skill has already been assessed. 
The interesting part is the set of unassessed skills of a learner with still unknown knowledge state, i.e., 
\begin{align*}
    \Ucal = \{s \in \Scal \,|\, \alpha(s) = 0 \} \subset \alpha(\Scal),
\end{align*} 
for which the model provides a probability for the knowledge of a learner.
We can eventually turn this probability into a binary prediction using a simple thresholding function $\phi_\tau \colon [0, 1] \to \{0, 1\}$ with parameter $\tau \in (0, 1)$,
\begin{align*}
    \phi_\tau(x) = 
    \begin{cases}
        0, & \text{if } x \leq \tau, \\
        1, & \text{if } x > \tau,
    \end{cases}
\end{align*}
such that, for any unassessed skill $s \in \Ucal$,
\begin{align*}
    \phi_\tau(\pi(s)) \in \{0, 1\}
\end{align*}
is a binary prediction for the learner's knowledge.
\begin{figure}
    \begin{center}
        \begin{tabular}{c}
            \newcommand{\inputnum}{5} 
\newcommand{\hiddennum}{4}  
\newcommand{\hiddennumtwo}{4}
\newcommand{\outputnum}{5} 

\begin{tikzpicture}

\foreach \i in {1,...,\inputnum}
{
	\node[circle, 
		minimum size = 6mm,
		fill=orange!30] (Input-\i) at (0,-\i) {};
}

\foreach \i in {1,...,\hiddennum}
{
	\node[circle, 
		minimum size = 6mm,
		fill=teal!50,
		yshift=(\hiddennum-\inputnum)*5 mm
	] (Hidden-\i) at (2.5,-\i) {};
}

\foreach \i in {1,...,\hiddennumtwo}
{
	\node[circle, 
		minimum size = 6mm,
		fill=teal!50,
		yshift=(\hiddennumtwo-\inputnum)*5 mm
	] (Hiddentwo-\i) at (5,-\i) {};
}

\foreach \i in {1,...,\outputnum}
{
	\node[circle, 
		minimum size = 6mm,
		fill=purple!50,
		yshift=(\outputnum-\inputnum)*5 mm
	] (Output-\i) at (7.5,-\i) {};
}

\foreach \i in {1,...,\inputnum}
{
	\foreach \j in {1,...,\hiddennum}
	{
		\draw[->, shorten >=1pt] (Input-\i) -- (Hidden-\j);	
	}
}

\foreach \i in {1,...,\hiddennum}
{
	\foreach \j in {1,...,\hiddennumtwo}
	{
		\draw[->, shorten >=1pt] (Hidden-\i) -- (Hiddentwo-\j);	
	}
}

\foreach \i in {1,...,\hiddennumtwo}
{
	\foreach \j in {1,...,\outputnum}
	{
		\draw[->, shorten >=1pt] (Hiddentwo-\i) -- (Output-\j);
	}
}

\foreach \i in {1,...,\inputnum}
{            
	\draw[<-, shorten >=1pt] (Input-\i) -- ++(-1,0)
		node[left]{$\alpha(s_{\i})$};
}

\foreach \i in {1,...,\outputnum}
{            
	\draw[->, shorten >=1pt] (Output-\i) -- ++(1,0)
		node[right]{$\pi(s_{\i})$};
}

\end{tikzpicture}
        \end{tabular}
    \end{center}
    \caption[Knowledge state network]{An exemplary, fully-connected knowledge state network for five skills with two hidden layers.}
    \label{fig:knowledge_state_network}
\end{figure}
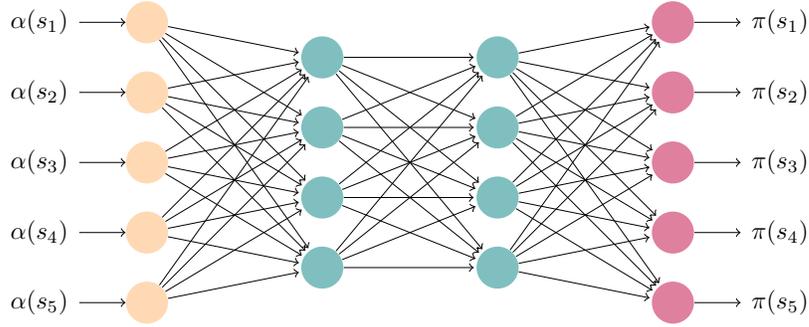
Since the input to the model -- the assessment state -- may only take values in a discrete set (skill unmastered: $-1$, skill mastered: $1$, skill state unknown $0$), our model will essentially behave like a ``decision gate'': particular combinations of input skill states will make output states more or less likely.
Note that this highly resembles the standard approach a teacher would take when trying to figure out the knowledge state of a student.

A special characteristic of the model is that, in contrast to many common prediction models, the output of the model is of the same dimension as the input. 
Naturally, over the course of the assessment, some part of the model output becomes redundant since the knowledge state of the respective skills is already known through a direct assessment question, such that the prediction does not provide any further value for these skills (or is potentially even wrong).
If one wants to avoid redundant output, a remedy can be to create a separate model for each skill $s \in \Scal$, taking an assessment state $A \in \Acal$ and outputting the probability of $s$ under this particular configuration. 
In view of a potentially very large number of skills to be assessed, however, this does not appear to be a viable solution.
In particular, we aim to reduce the amount of assessment steps to a number much smaller than the actual number of skills, such that the redundancy of outputs stays rather insignificant in practice.

Model-wise, our weapon of choice is a relatively standard feed-forward neural network, consisting of multiple fully-connected layers with the rectified linear unit as the activation function (cf. Figure \ref{fig:knowledge_state_network}). 
In the same fashion as for standard classification models, the output layer of our model features the sigmoid function 
\begin{align*}
    \mathrm{sig}(x) = \frac{1}{1 + e^{-x}}
\end{align*}  
as the activation function of the output layer in order to ``squeeze'' each of the outputs between $0$ and $1$, hence resembling a probability.

The training objective of the model, in contrast to standard classification, is chosen to be the mean squared distance between the output of the model and the true knowledge states in $\{0, 1\}$, i.e., for any assessment state $A \in \Acal$,
\begin{align}
    \frac{1}{|\Scal|} \| \Mcal(A) - \sigma(\Scal) \|^2 
    = \frac{1}{|\Scal|} \sum_{s \in \Scal} |\pi_\Mcal(s) - \sigma(s)|^2 \label{eq:training_objective}
\end{align}
We hence obtain a model which tries to match the output as close as possible to the binary true assessment.

\subsection{Simulating assessment states to train the model}\label{subsec:knowledge_state_simulation}
One of the most interesting questions is how to generate appropriate training data for the model. 
It essentially requires to see different assessment states, i.e., incomplete knowledge states as the input and has to predict the full, true knowledge state, making the full, true knowledge state our target. 
The data we have at hand are full knowledge states of previous learners that have already gone through the assessment.
These build the target states for our model to train, and can as well be used to sample incomplete assessment states from them by simply covering a random number of knowledge states. 

However, recalling the amount of possible assessment states $|\Acal| = 3^n$, a sufficiently dense sampling seems to be increasingly impossible with a growing number $n$ of skills.  
The key here is that, even though the cardinality of $\Acal$ is huge, the majority of possible knowledge states is entirely impossible to achieve for learners in practice and is hence irrelevant for our model. 
For instance, imagine a learning path with one hundred skills from a certain topic that need to be assessed.
Due to the inherent, logical connection between skills, some skills simply cannot be mastered before others have been mastered. 
Simply speaking, a learner most likely will not have mastered the concept of a for-loop before having understood the idea of a variable when studying a programming language. 
This, together with the naturally increasing difficulty of most topics, leads to learner knowledge typically concentrated at the start of a learning path, decreasing towards more advanced topics (see also Figures \ref{fig:user_path_assessment_mastery} and \ref{fig:user_assessment_mastery_outlook} for typical knowledge distributions in learning paths).

\begin{algorithm}[t]
    \caption{Assessment state simulation}\label{pseudo:assessment_state_simulation}
    \begin{algorithmic}[1]
        \State \textbf{Input}: learners of the training set, $k$ number of simulations per learner 
        \For{each learner $i$ and $j = 1, \dots, k$}
            \State \textbf{initialize} state $A_j$ with $\alpha(s) = 0$ for all $s$ 
            \State \textbf{choose} a random number $m \leq n$ 
            \State \textbf{choose} a subset $S_m \subset \Scal$ of $m$ random skills  
            \For{each $s \in S_m$}
                \State \textbf{set} $\alpha(s) = 1$ if $\sigma_i(s) = 1$
                \State \textbf{set} $\alpha(s) = -1$ if $\sigma_i(s) = 0$
            \EndFor
            \State \textbf{return} assessment state $A_j$
        \EndFor
    \end{algorithmic}
\end{algorithm}

Algorithm \ref{pseudo:assessment_state_simulation} lines out a procedure to sample \textit{possible}, incomplete assessment states from a learner, Table \ref{tab:assessment_state_simulation} shows a few possible and impossible samplings for our simple ontology in Figure \ref{fig:ontology_states}. 
It is important to notice that the algorithm only creates assessment states that could have occurred during the assessment of the learner; an assessment state with, e.g., $\alpha(s_1) = -1$ would not be possible since the learner had already mastered $s_1$ before the assessment (i.e., $\sigma(s_1) = 1$).

Yet another way to see that the procedure indeed produces the correct, incomplete assessment states for training is the model itself. 
As in almost every situation, machine learning models, while performing great at interpolation, are inherently bad at extrapolation.
Simply put, we can expect our model to capture the knowledge structure of learner types that it has seen in training, but we cannot expect it to predict useful probabilities for learner types it has never encountered. 
Training on learners with little to no knowledge only will not result in a useful model for experienced learners and vice versa. 
However, using data from different learner types for training will create a useful model also for different situations.\footnote{
    Of course, one can also try to figure out the ``learner type'' prior to the assessment and use a different model depending on the type.}

These considerations underline that, in practice, the relevant part of assessment states is of a significantly smaller complexity than $|\Acal|$ such that our learner sampling represents a viable approach.
It is worth mentioning that we can also include the knowledge states of previous learners that have not gone through all of the skills in $\Scal$ during their assessment, e.g., due to session learning (see Section \ref{sec:assessment_strategies}) such that we have an incomplete target state. 
In this case we set $\sigma(s) = 0.5$, indicating that the true knowledge state is unknown also during training\footnote{Even tough $0.5$ is technically not included in our definition of $\sigma$.}.
Of course, this is only viable for a certain percentage of missing knowledge states since otherwise the training becomes erroneous.
In practice, we use learners for training where at least eighty percent of the true knowledge state is known.

\begin{figure}
    \begin{center}
        \resizebox{0.55\textwidth}{!}{
            \begin{tabular}{ l | c c c c c c}
                Skill & $s_1$ & $s_2$ & $s_3$ & $s_4$ & $s_5$ & $s_6$ \\[2pt]
                \hline
                \hline\\[-5pt]
                true $\sigma(s)$ & $1$ & $1$ & $1$ & $0$ & $1$ & $0$ \\[2pt]
                \hline\\[-5pt]
                possible simulated $\alpha(s)$ & $1$ & $1$ & $0$ & $0$ & $0$ & $0$ \\
                & $1$ & $0$ & $0$ & $-1$ & $1$ & $0$ \\
                & $1$ & $0$ & $1$ & $0$ & $0$ & $0$ \\
                & $1$ & $1$ & $1$ & $0$ & $0$ & $-1$ \\
                & $0$ & $0$ & $0$ & $0$ & $1$ & $0$ \\
                & & & $\dots$ \\[2pt]
                \hline\\[-5pt]
                impossible $\alpha(s)$ & $-1$ & $-1$ & $0$ & $0$ & $0$ & $0$ \\
                & $1$ & $-1$ & $0$ & $0$ & $0$ & $1$ \\
                & & & $\dots$ \\[2pt]
            \end{tabular}
        }
    \end{center}
    \caption{Possible and impossible assessment state simulations/samplings from a learner with knowledge state $\sigma(\Scal)$.
    Note, that only assessments states can be sampled, which fit the true knowledge state of the learner.}
    \label{tab:assessment_state_simulation}
\end{figure}

\subsection{Maintaining the training data and solving the cold start problem}
The goal of using such models in practice is to decrease the amount of time spent on skill assessment to save additional time for learning by predicting the knowledge state of skills rather than directly assessing them. 
However, it is of high importance to get the learner's feedback for the prediction. 
The main two reasons are the learner's satisfaction and the maintenance of the training data base. 
The former is easily explained: 
Due to the heterogeneity of learners in general and the inevitable errors of the prediction model it can happen that the model predicts skills as known that are actually unknown and vice versa. 
This can lead to a bad learning experience, when necessary skills are skipped by mistake such that subsequent skills cannot be understood by the learner, or skills that are already known repeatedly show up in the learning process despite the learner's knowledge. 
We hence need to empower the user to correct the prediction of our model in order to guarantee a smooth and satisfying learning experience.

The second reason for explicit user feedback about the performance of our model is that, without it, we slowly but steadily corrupt the training basis for our model, such that the model does not get better but worse with every new learner going through the assessment.
For the same reasons as above we cannot use the model predictions as target states for our model training since they inevitably contain errors which degrade our model's quality. 
We hence need to make sure that the learner has the possibility to either confirm or correct the predictions of our model.
At edyoucated, this process is directly integrated into the learning flow and does not impose any obstacle for the learner. 

An obvious obstacle, however, is the ``cold start'' problem. 
Naturally, as long as no learner has gone through the assessment yet, we do not have any assessment data to train the neural network on.
We like to line out three different possibilities to bypass the standard knowledge state network until a critical mass of training data has been acquired. 
In view of Section \ref{subsec:atomic_ontologies}, an obvious idea is to use the concept of prerequisites (cf. also \cite{Falmagne2006}) to speed up the assessment process which, as argued, is not a convenient option for long learning paths. 
Another obvious approach is to simply gather a batch of learners who are willing to take the full assessment, i.e., who answer every single assessment question in one single sitting in order to gather the necessary training data.
As volunteers are sometimes hard to find, an easier and more viable option in most cases is to \textit{define} a few reasonable, full assessment states for heterogeneous learner types and to feed those to the network to learn, starting with very little knowledge up to a higher amount of mastery. 
Of course, the results will not be as sophisticated as the results for a high amount of assessment data from real learners, but can serve as a good approximation to get started.
In any of the above strategies, we eventually use a knowledge state network once a critical mass of true assessment states has been acquired. 
In our experience, it is often already sufficient to have more than ten to twenty full assessment states (respectively, knowledge states) to achieve good results with a network. 
Note, however, that this also depends on the heterogeneity of the assessed learners; if only similar learner types go through the assessment we might need to wait a little longer to shift to the use of a trained network.

\subsection{Evaluating the model}
The evaluation of the model is done using standard routines from machine learning. 
We investigate the general performance of the model in the standard situation of training and evaluation on a training set and a test set of learners, where for all learners the true knowledge state is known.
We use three different measures in order to quantify the performance of our model, where one can also be used as a stopping criterion in practice. 

In view of the model's training objective \eqref{eq:training_objective}, the most straightforward way to measure the error of the model is to compute either the squared or absolute error between the true knowledge state and the knowledge probabilities given by the model on an evaluation set. 
In accordance with the model training we use the \textit{mean squared probability error}, 
\begin{align*}
    \mathrm{MSPE}(\pi, \Scal) = \frac{1}{|\Scal|} \sum_{s \in \Scal} |\pi(s) - \sigma(s)|^2.
\end{align*}
We define the \textit{relative knowledge state error} as a more ``binary'' version of the error using the thresholding function $\phi_\tau$,
\begin{align}
    \mathrm{RKSE}_\tau(\pi, \Scal) = \frac{1}{|\Scal|} \sum_{s \in \Scal} |\phi_\tau(\pi(s)) - \sigma(s)|.
    \label{err:relative_knowledge_state_error}
\end{align}
Depending on the threshold $\tau$, it captures the average error we make when letting the model predict the knowledge state of the learner.
Note that both of the above errors may also be evaluated only for the set $\Ucal$ of unassessed skills (see also Section \ref{sec:numerics}). 
This offers the advantage that the error is only evaluated on skills where we do not know the true knowledge state yet such that, over the course of the assessment, the error does not decay only by virtue of asking more questions and receiving correct answers but by the network predictions getting more precise. 
The last measure we would like to introduce is the \textit{knowledge state uncertainty}, which expresses the uncertainty in the model when doing a prediction. 
Naturally, the closer the probability $\pi(s)$ of a skill $s$ is to 0 or 1, the more certain one can be with the eventual prediction of the model for the skills state of $s$.
Hence, for a small parameter $\eps$, we define the knowledge state uncertainty error as the cardinality of the set of uncertain skills, i.e.,
\begin{align}
    \mathrm{KSUE}_\eps(\pi, \Scal) = | \{s \in \Scal \,|\, \eps \leq \pi(s) \leq 1 - \eps\} |.
    \label{err:knowledge_state_uncertainty}
\end{align}
The latter measure can also be used to stop the assessment once a sufficiently low uncertainty has been reached, since it does not require any knowledge about the true knowledge state of the learner. 
Clearly, using a naive assessment with no connections between skills, we obtain a linear decrease of \eqref{err:knowledge_state_uncertainty}, one at a time. 
A more involved model hence aims at achieving a much steeper decrease in \eqref{err:knowledge_state_uncertainty} such that information about the knowledge state of one additional skill causes the uncertainty about multiple other skills to drop below the threshold.

\section{Assessment strategies}\label{sec:assessment_strategies}
With our knowledge state network as a predictor at hand, we can start designing our assessment strategies. 
While there are many different approaches to design an assessment, we want to stick to two different classes of assessments here: 
full assessments and session assessments. 
The task for the former consists in assessing the learner's knowledge state for a full set $\Scal = \{s_1, s_2, \dots, s_n\}$ of $n$ skills in a single session. 
Using standard assessment methods, this can take a substantial amount of time for a larger amount of skills.
It is worth noting that the assessment here does not necessarily have to be ordered since, by the end of the assessment, we will have full information about the knowledge state and can then assemble the unknown skills in the correct learning order to create a learning path for the learner.

For session learning, the idea is to split the task of learning all skills of $\Scal$ into $m$ disjoint learning sessions $L_1, \dots, L_m$, such that $L_i \cap L_{j} = \emptyset$ for all $i \neq j$, each containing a few next skills to learn. 
Once a learning session $L_i$ has been finished, the learner can proceed to the next session $L_{i+1}$.
Between each two learning sessions, however, the learner will have a short assessment in order to determine the content of the next learning session.
For the assessment of session $L_i$, the goal is hence to find the next $l_i$ learnable skills in $\Scal$, i.e., the next $l_i$ skills that are unknown to the learner, in order to create a new learning session.
Mathematically speaking, we need to find the following set of learnable skills, 
\begin{align}
    \Scal_{\mathrm{next}} = \{s_j \in \Scal \,|\, &\alpha(s_j) = -1, \alpha(s_m) \neq 0 \,\forall\, m \leq j \},\label{eq:next_learnable_skills}
\end{align}
and then take the first $k$ skills from that set,
\begin{align}
    \Scal_{l_i} = \{s_j \in \Scal_{\mathrm{next}} \,|\, 1 \leq j \leq l_i \}.
    \label{eq:subs_next_learnable_skills}
\end{align}
The difficulty here is that the skills in $\Scal$ are necessarily ordered, and we have to guarantee that we do not leave any potential knowledge gaps for the learner, i.e., skills $s$ where $\alpha(s) = 0$ and we do not know, yet, whether the learner has mastered $s$ or not. 

The general goal for both assessment variants is to ask as few questions as possible while, of course, the assessment for a session should be very short for a small number of session skills, while for larger sessions also the assessment can be longer. 
In case of a full assessment, we stop the assessment once the knowledge state uncertainty error \eqref{err:knowledge_state_uncertainty} reaches zero, implying that the model is sufficiently certain about the full knowledge state of the learner.
For session learning we can do the same, however, evaluating the uncertainty on the set of next learnable skills only.
Without loss of generality we assume that we do not have any knowledge about the learner at the start of the assessment, i.e., $\alpha(s) = 0$ for all $s \in \Scal$. 
Of course, in practice, additional prior knowledge about some skills of the learner can be integrated into the model in the same fashion or, if we are already aware of the learner's knowledge state for some skills we can obviously include this as a starting point. 
The assessment strategy, however, does not change significantly, so we assume no prior information about the learner's knowledge state in the following.

\subsection{Picking the next skill for the assessment}
There are numerous ways to choose which skill should be assessed next, and this highly depends on the application and the related user experience as well as the assessment goal. 
Often, the mathematically best next question, in some sense, might not be the next ``logical'' question when it comes to the inherent order of the skills in $\Scal$. 
Simply speaking, an automatically chosen question might differ from what we would expect, e.g., a teacher to ask next. 
In the following, we line out three different strategies that have been proven effective, which can be adapted to the particular assessment situation at hand.
The goal, as already pointed out above, is to keep the assessment as short as possible, i.e., to reach a knowledge state uncertainty error \eqref{err:knowledge_state_uncertainty} of zero as quickly as possible.

The easiest and often quite successful approach to ask questions from a mathematical perspective is to simply choose questions at random. 
The obvious downside is the presumably bad user experience if the assessment jumps through different topics in no apparently ``logical'' order. 
A better approach is to start trying to reduce uncertainty. 
A simple heuristic is to always assess a skill $\sopt$ with the highest residual uncertainty in the model, i.e., a skill with probability closest to $0.5$,  
\begin{align*}
    \sopt \in \arg \min_{s \in \Scal} |\pi(s) - 0.5|.
\end{align*}  
Intuitively, a skill with probability close to $0.5$ is likely a skill that roughly half of the learners from the training set have marked as known respectively unknown (given the current assessment state), such that it allows to essentially split the amount of learners in half in order to find similar learner types. 
Hence, such an approach is likely to pick rather critical skills in order to determine a learner's knowledge.

\begin{figure*}[!t]
    \centering
    \captionsetup[subfigure]{
        labelformat=empty, 
        justification=centering
    }   
    \begin{subfigure}[c]{0.49\textwidth}
        \centering
        \includegraphics[height=0.28\textheight]{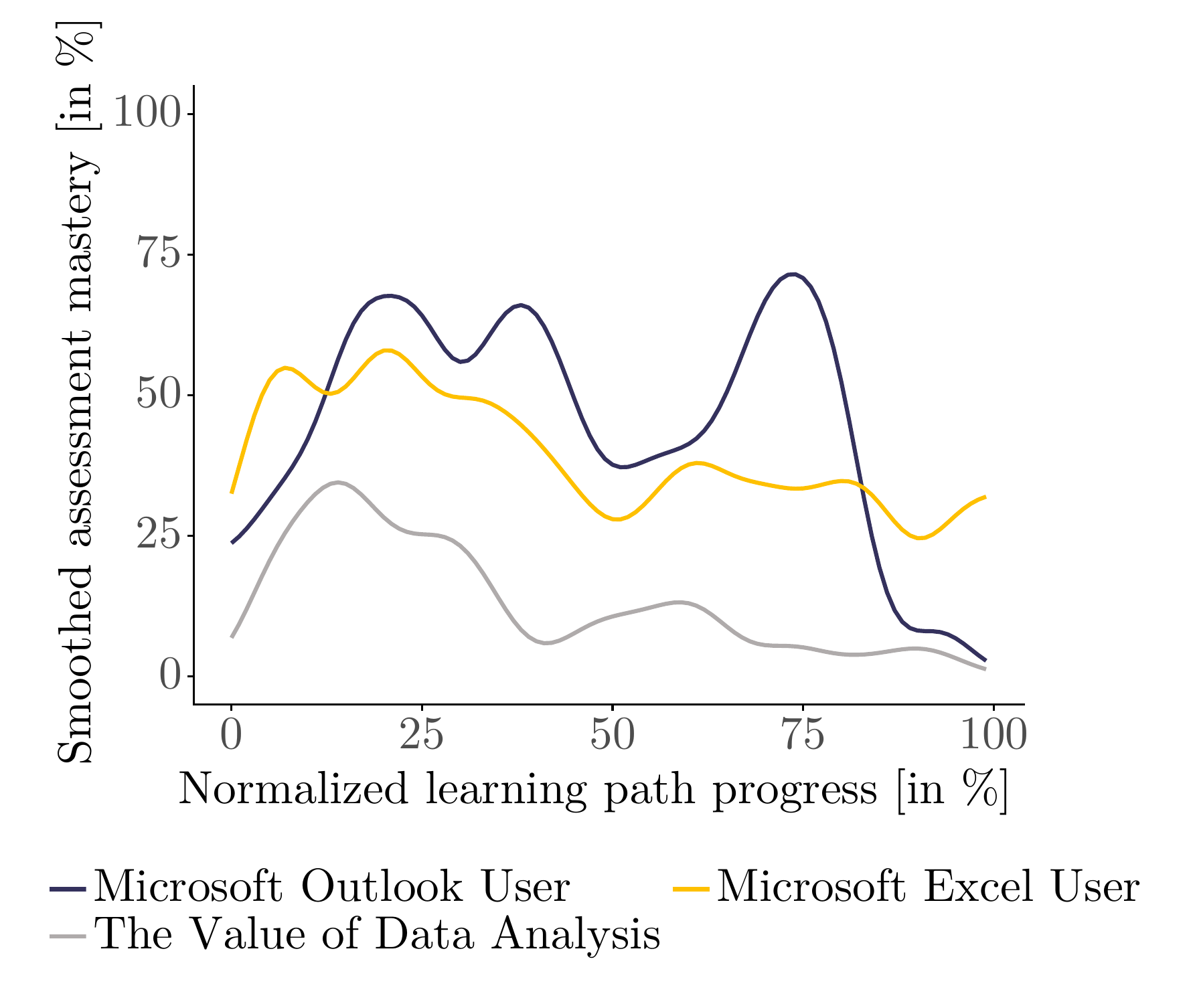}
        \subcaption{a) (Smoothed) assessment mastery for different learning paths}
        \label{fig:user_path_assessment_mastery}
    \end{subfigure}     
    \begin{subfigure}[c]{0.49\textwidth}
        \centering
        \includegraphics[height=0.28\textheight]{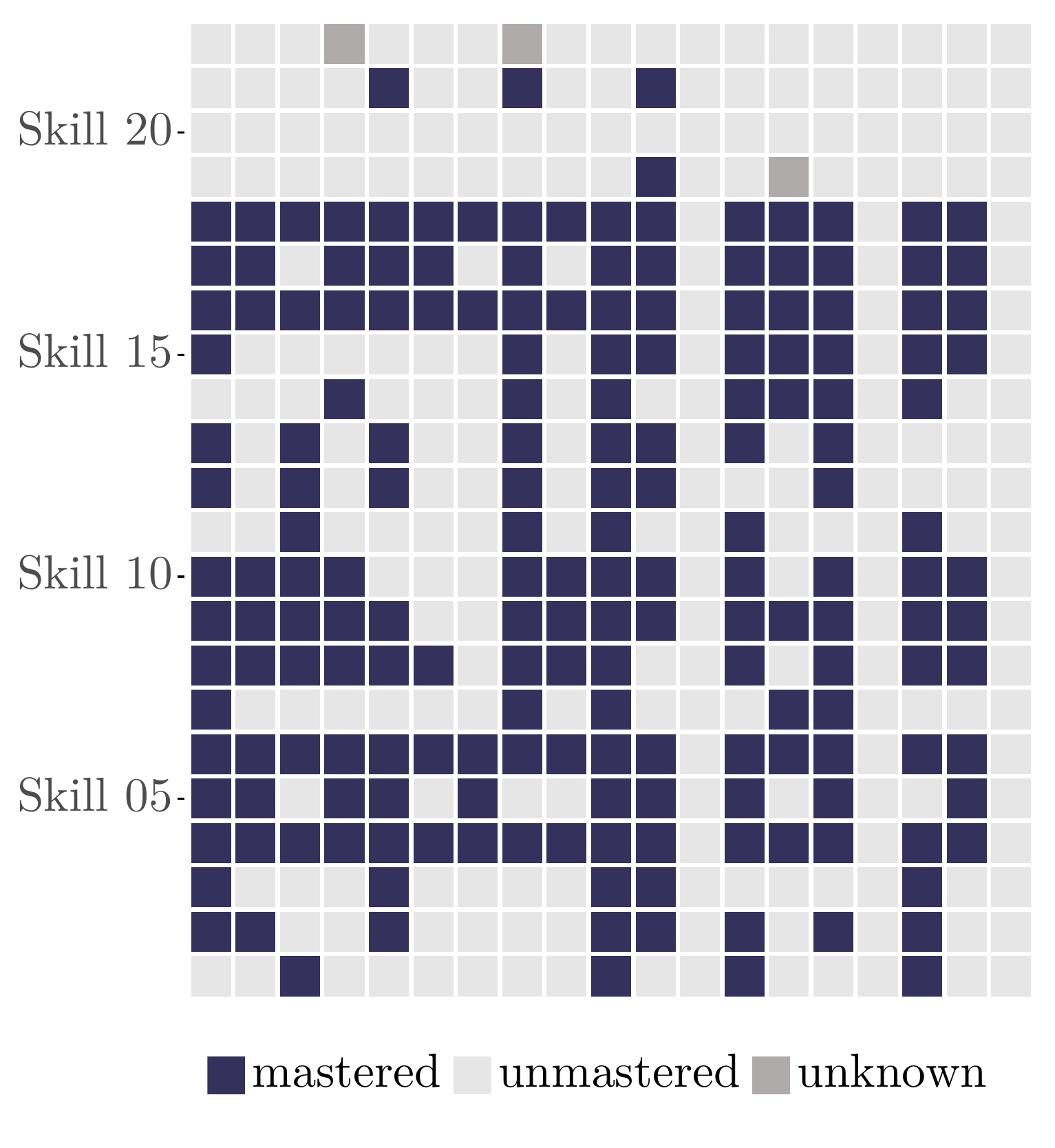}
        \subcaption{b) Detailed learner assessment mastery for \textit{Microsoft Outlook}}
        \label{fig:user_assessment_mastery_outlook}
    \end{subfigure} 
    \caption{
        (a) (Smoothed) average assessment mastery for different learning paths on the edyoucated platform.
        For some topics, prior knowledge concentrates rather in the beginning, for some it is spread out over the learning path.
        Learning path length -- they can have different amounts of skills -- has been normalized for all paths.
        (b) Learner assessment mastery for the learning path \textit{Microsoft Outlook}.
        The columns of the map indicate the different learners.
        Unknown skill mastery occurs when learners have not fully completed the assessment on the platform.
    }
    
\end{figure*}

Of course, one can take the idea of reducing uncertainty a step further and try to find the skill $\sopt$ causing the optimal decrease in the uncertainty error \eqref{err:knowledge_state_uncertainty} in each step of the algorithm.
Due to the discrete nature of our assessment (only one skill can change its assessment state at a time), this boils down to a combinatorial problem where we essentially have to compute the value of \eqref{err:knowledge_state_uncertainty} for each possible skill we can assess.
There remains, however, uncertainty about the answer of the learner. 
While the information that some skill $s$ is known might decrease our objective substantially, it can occur that, if the skill is unknown, we do not decrease our objective at all. 
A remedy can be found, again, in the likelihood of a learner answering positive or negative provided by our model, in order to weight the positive and negative answers.
We hence propose to use the optimal \textit{expected} decrease in uncertainty by minimizing the knowledge state uncertainty error \eqref{err:knowledge_state_uncertainty} weighted by the probabilities of positive or negative transition between skills assessment states. 

More precisely, for two consecutive assessment steps $t_k$ and $t_{k+1}$, let $\pi_s^+(\cdot, t_{k+1})$ be the probability function resulting from the learner answering the assessment for skill $s$ as known, i.e., $\alpha(s, t_{k+1}) = 1$. 
Similarly, let $\pi_s^-(\cdot, t_{k+1})$ be the probability function emerging for a negative answer for $s$ ($s$ unknown).
Then, for each skill $s \in \Scal$, we can define the expected uncertainty as
\begin{align*}
    \Delta(\pi, s) &= \pi(s, t_k) \ \mathrm{KSUE}_\eps(\pi_s^+( \cdot, t_{k+1}), \Scal) + (1-\pi(s, t_k)) \ \mathrm{KSUE}_\eps(\pi_s^-( \cdot, t_{k+1}), \Scal). 
\end{align*}
A skill causing the optimal decrease in uncertainty can then be found solving 
\begin{align*}
    \sopt \in \arg \min_{s \in \Scal} \Delta(\pi, s).
\end{align*}
A small modification one can make in practice is to replace the ``discrete'' knowledge state uncertainty error \eqref{err:knowledge_state_uncertainty} in the above formula by a more precise version, measuring the residual uncertainty in terms of probabilities, 
\begin{align*}
    \delta(\pi, \Scal) = \dfrac{1}{|\Scal|} \sum_{s \in \Scal} \min \{ \pi(s), 1 - \pi(s) \}.
\end{align*}
Note that, to speed up the process in practice, we can evaluate the above measures for the set $\Ucal$ of unknown skills only.
It is clear, that the entire strategy is rather local in the sense that it cannot guarantee a globally optimal decrease in uncertainty. 
If the amount of skills to be assessed is rather large, one can additionally employ a sampling strategy and only evaluate the expected descent for a random number of skills to choose from.
Of course, the approach can as well be extended to multiple steps, simulating the next few iterations of the assessment in order to find the best suitable descent path in expectation. 
The issue here is the increasing complexity of the simulation since the number of objective values we need to compute grow exponentially.
A potential remedy here, again, is the sampling of simulation paths and choosing among these which, however, is subject to future work.

In practice, a combination of the outlined strategies has proven to be effective: 
among the most uncertain skills (say, between 5 and 10) we choose the one promising the steepest expected descent with respect to uncertainty. 
This both keeps the approach computationally viable and contributes well to the eventual triggering of our stopping criterion.

\subsection{Session learning: Exploration vs. confirmation}
All the strategies mentioned above can be employed in order to accelerate assessment processes for learners, and are particularly valuable for full assessments of a larger number of skills.
We provide a practical study and comparison for full assessments in Section \ref{sec:numerics}.

A potential drawback of the above assessment methods is that the assessment order will most probably not follow the learn order of the skills. 
Indeed, to guarantee a steep descent of our objective and eventually a relatively short assessment, the skills that are assessed (and their order) usually have to be quite spread out over the whole learning path in order to predict the full knowledge state of a learner with sufficient confidence.
Hence, for topics with an inherent learn order, there is no guarantee that the algorithms provide a subsequent set of skills that the learner can learn \textit{before} the assessment has been entirely completed.
For a large number of skills it may still require a relatively large amount of skills to be assessed.
This can be prohibitive for the motivation of the learner who, in general, wants to start learning as quickly as possible instead of spending a substantial amount of time on an assessment. 

The key idea to solve this issue is session learning, which requires to tailor our algorithms to work in such a setting.
As lined out at the beginning of this section, session learning consists of finding the next $l$ learnable skills from an \textit{ordered} set of skills $\Scal$, i.e., the next $l$ skills that are unknown to the learner, without leaving any skill gaps in between. 
The learner can then directly proceed with learning these $l$ skills before continuing the assessment in order to create the next session. 
This approach alternates the less interesting assessment part with the (hopefully) more interesting learning part. 

The goal is now twofold: 
On the one hand, we want to achieve a general decrease in uncertainty about the learners knowledge state for the entirety of the skills $\Scal$ (eventually these skills will appear in one session and we hence need to gather the information anyway). 
On the other hand, we want to guarantee that we indeed have the $l$ next learnable skills at hand after only a few questions in order to keep the assessment sessions as short as possible. 
To achieve a balancing of these two goals, we propose to use a classic \textit{exploration vs. confirmation} strategy. 
For each step of the assessment, we employ a probability $p \in [0, 1]$ which balances exploration, with the aim of a global decrease of uncertainty about the learner's knowledge state, and confirmation, restricting the assessment process to the quick discovery of the $l$ next learnable skills. 
With a probability of $p$ we use the above assessment strategies in order to find the next assessed skill among \textit{all} $s \in \Scal$, with probability $1-p$ we use a strategy to find the next assessed skill among the next $l$ learnable skills $\Scal_\mathrm{next}$ (cf. \eqref{eq:next_learnable_skills}).
Depending on the choice of $p$, this allows to ask for the globally most discriminative skills during the assessment while ensuring that we also gain knowledge about the very next skills the learner has to learn.

\begin{algorithm}
    \caption{Exploration vs. confirmation}\label{pseudo:exploration_vs_confirmation}
    \begin{algorithmic}[1]
        \State \textbf{Input}: exploration prob. $p$, session length $l$ 
        \State \textbf{determine} next learnable skills $\nls^{(\pi)}$
        \State with probability $p$ \textbf{choose} exploration
        \If{not exploration}
            \State \textbf{choose} subset $\Scal_l^{(\pi)}$ of $\nls^{(\pi)}$ with $|\Scal_l^{(\pi)}| = l$
            \State \textbf{choose} $s$ from $\Scal_l^{(\pi)}$ with any strategy
        \Else
            \State \textbf{choose} $s$ from $\nls^{(\pi)}$ with any strategy
        \EndIf 
    \end{algorithmic}
\end{algorithm}

In order to make the strategy even more viable in practice, we can furthermore relax the conditions on the set of next learnable skills \eqref{eq:next_learnable_skills}, again using the prediction of our knowledge state network. 
The set of \textit{predicted} next learnable skills is 
\begin{align*}
    \Scal_{\mathrm{next}}^\pi = \{s_j \in \Scal \,|\, &\pi(s_j) \leq \eps, \pi(s_m) \notin (\eps, 1- \eps) \,\forall\, m \leq j \}.
\end{align*}
Intuitively, this is the set of all skills such that the probability of the learner knowing a skill is below a (small) threshold $\eps$, and all skills between such skills need to have a knowledge probability of at least $1- \eps$, i.e., are most likely known and can be skipped.
Recall, that for every $s$ such that $\alpha(s) = -1$ also $\pi(s) = 0$.
The predicted learnable skills may be used also as a stopping criterion for session assessment; the assessment can be stopped once $\Scal_l^\pi$ has been identified (cf. \eqref{eq:subs_next_learnable_skills}).

\section{Numerical studies}\label{sec:numerics}
\begin{figure}[t!]
    \centering
    \includegraphics[width=\textwidth]{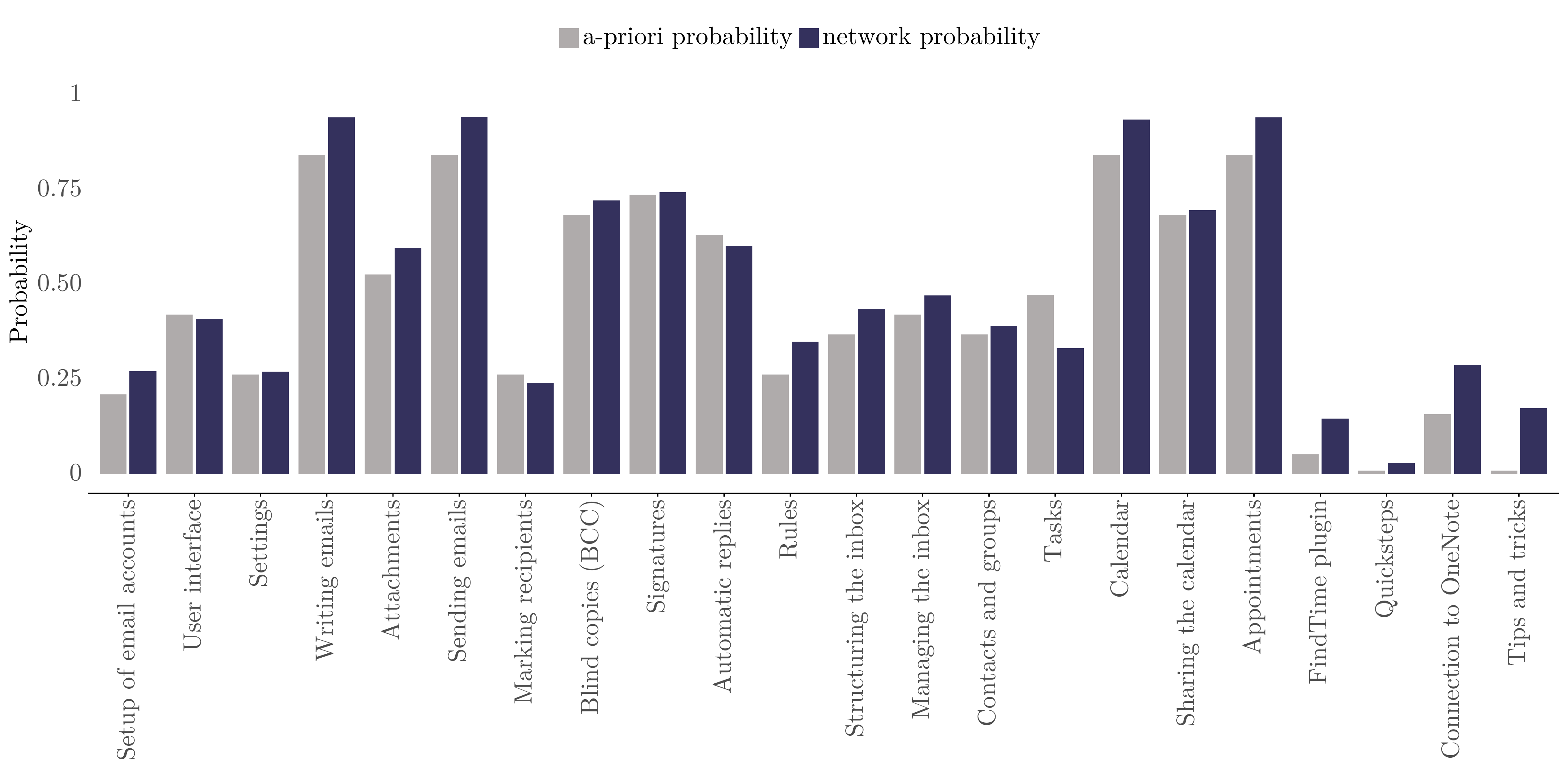}
    \caption{
        A-priori mastery probability (percentage of learners that mastered the skill) vs. learned a-priori network probabilities (for an entirely incomplete knowledge state) for the learning path \textit{Microsoft Outlook}.
    }
    \label{fig:initial_probabilities_outlook}
\end{figure}

In this section we want to provide numerical proof of concepts for the ideas and methods introduced above.
In order to demonstrate the performance of the algorithms we use pseudonymized learner data from the edyoucated learning platform and ``re-simulate'' the assessment process for the learners in different settings. 
More precisely, the data we use is assessment data for three different learning paths offered on the platform based on an atomic skill ontology. 

Before we go into detail about the respective paths we want to line out \textit{why} we chose these particular ones. 
As we will see throughout the experiments, they cover different ``assessment situations'', more precisely, different amounts and structure of average prior knowledge. 
Figure \ref{fig:user_path_assessment_mastery} sheds some first light on the distribution of assessment mastery
\footnote{We use the terms \textit{assessment mastery} and \textit{prior knowledge} interchangeably.}, i.e., the skills learners marked as mastered respectively unmastered during the assessment, 
over the different skills of the learning path. 
As discussed above, each path has an inherent order in which skills are (or should be) learned where, in the sense of personalization, of course parts can and will be skipped. 
However, it is interesting to see \textit{in which part} of the learning path prior knowledge is more likely to occur. 
A typical situations is that prior knowledge occurs towards the beginning of a path (i.e., the basics of a topic) and then decays as the topic gets more involved (cf. \textit{The Value of Data Analysis}). 
But there is as well the situation where skills do not depend too much on each other (or different topics are put together in a single path) such that prior knowledge is spread out over the entirety of the path (cf. \textit{Microsoft Outlook}). 
We hence need to ensure that our algorithms work for any amount of prior knowledge in any possible part of the learning path, and we will see that this is indeed the case. 
It is worth noting that prior knowledge often does \textit{not} peak on the first skills since they are not necessarily the easiest or most commonly known. 
For example, the first skills of a learning path could deal with the installation or setup of a tool which is then followed by its use. 
In many cases learners might not have gone through the setup process on their own since that process has been done, e.g., by their technology department, such that their prior knowledge ``starts'' a little later with the actual usage of a tool (cf. Figure \ref{fig:initial_probabilities_outlook} for such a situation). 

Let us have a look at the different learning paths in more detail.
The first learning path is named \textit{The Value of Data Analysis} and gives a general overview of the benefits and limitations of data analysis and related topics. 
It comprises a total number of 39 skills whose mastery needs to be assessed prior to the learning process.
As can be seen in Figure \ref{fig:user_path_assessment_mastery}, the mastery concentrates rather in the beginning of the learning path, where up to forty percent of the learners have mastered some skills; skills towards to the end of the learning path are predominantly unknown (see also the breakdown in Figure \ref{fig:user_assessment_mastery_the_value_of_data_analysis} of the Appendix).
The second learning path we investigate teaches the learner the skills of \textit{Microsoft Outlook}, i.e., the use of a fairly popular email and calendar communication tool.
As there is rather limited functionality that needs to be mastered about Outlook, the maximum number of learnable skills in this role is 22. 
Again, Figure \ref{fig:user_path_assessment_mastery} shows the assessment mastery of the learners. 
We observe that an average of roughly fifty percent of the learners have already mastered most of the skills, with a peak in mastery in the later part of the learning path. 
Figure \ref{fig:user_assessment_mastery_outlook} shows the assessment mastery of the learners in more detail, revealing some learners with little or extensive prior knowledge, and in particular a rather heterogeneous overall distribution of knowledge. 
The third and final learning path we investigate is \textit{Microsoft Excel}, comprising a total number of 57 atomic skills. 
Similar to the \textit{Microsoft Outlook} path it is particularly interesting due to the heterogeneous distribution of prior knowledge and mastery (see also Figure \ref{fig:user_path_assessment_mastery} and Figure \ref{fig:user_assessment_mastery_excel} of the Appendix).

We do not have any explicit demographic data about the learners themselves; however, since they all participated in an upskilling program, they are most likely aged 25-55 and share some interest in lifelong learning.
The inclusion of demographic data and other factors for further personalization are part of ongoing research.
For the simulation of the training data we followed the approach lined out in Section \ref{subsec:knowledge_state_simulation} and generated between one and ten thousand random knowledge states of each learner, roughly depending on the amount of skills in each learning path.

\begin{figure*}[!t]
    \centering
    \captionsetup[subfigure]{labelformat=empty, justification=centering}   
    \begin{subfigure}[c]{0.32\textwidth}
        \centering
        \includegraphics[width=\textwidth]{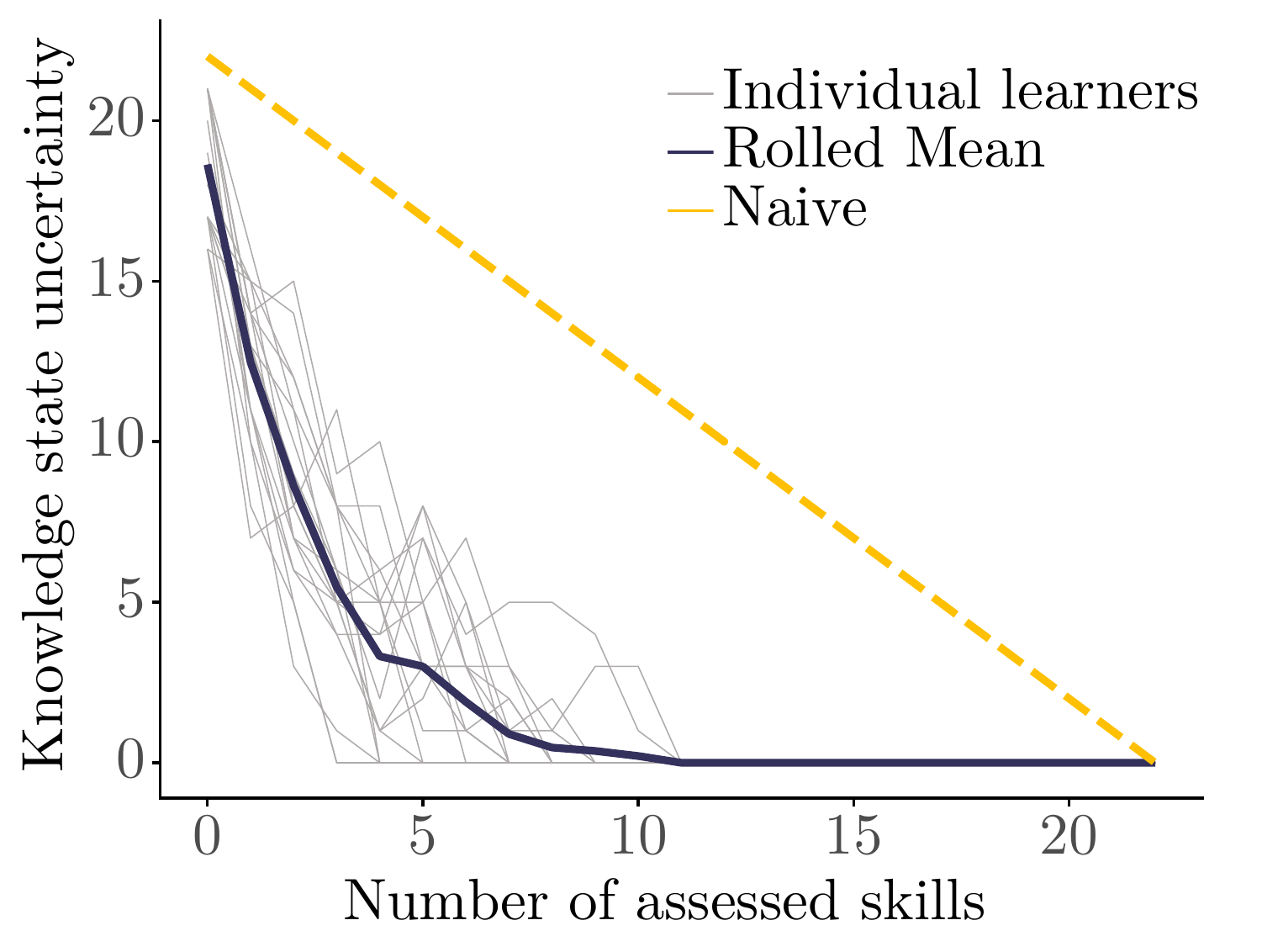}
        \subcaption{Microsoft Outlook}
    \end{subfigure}     
    \begin{subfigure}[c]{0.32\textwidth}
        \centering
        \includegraphics[width=\textwidth]{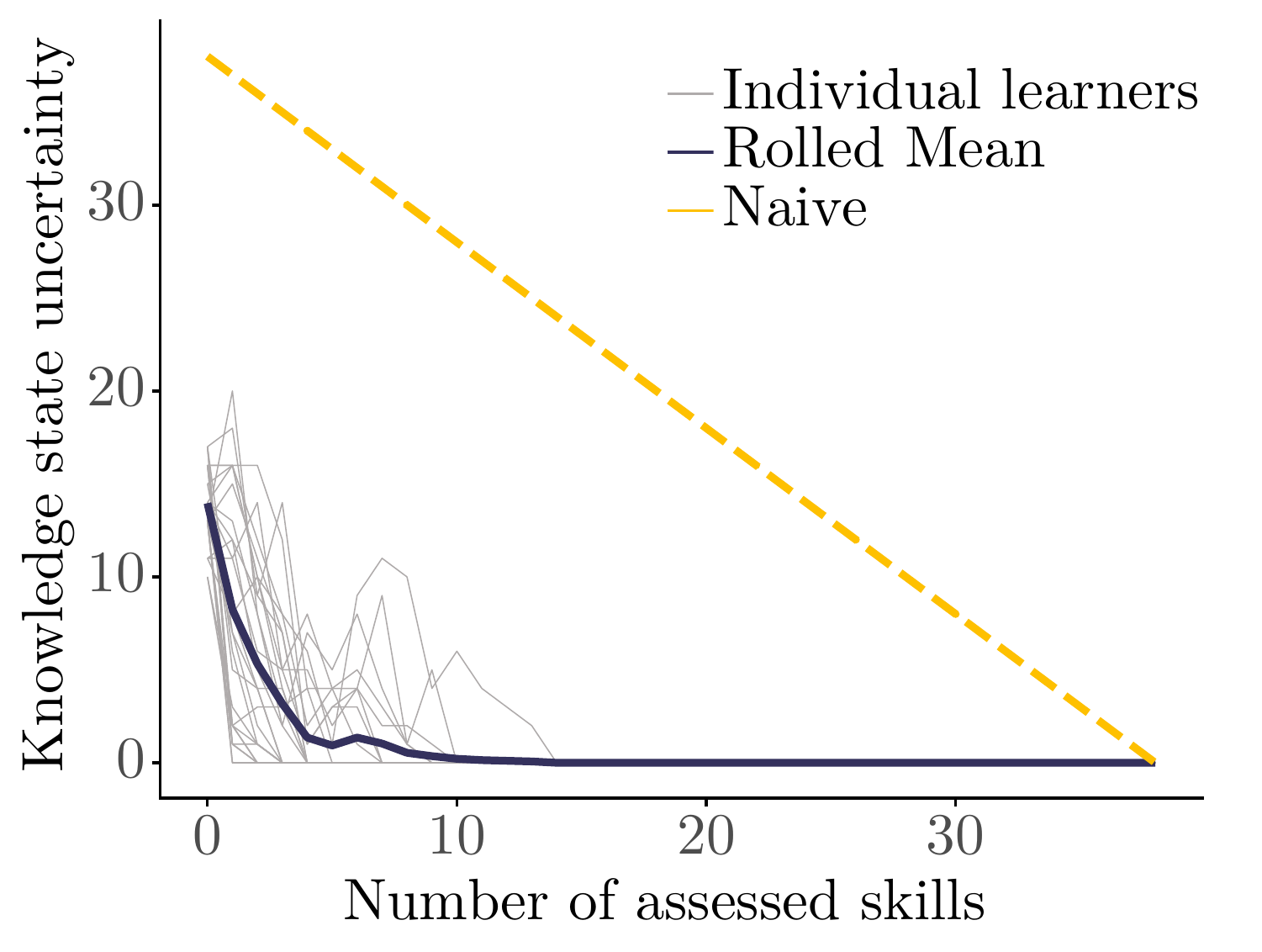}
        \subcaption{The Value of Data Analysis}
    \end{subfigure} 
    \begin{subfigure}[c]{0.32\textwidth}
        \centering
        \includegraphics[width=\textwidth]{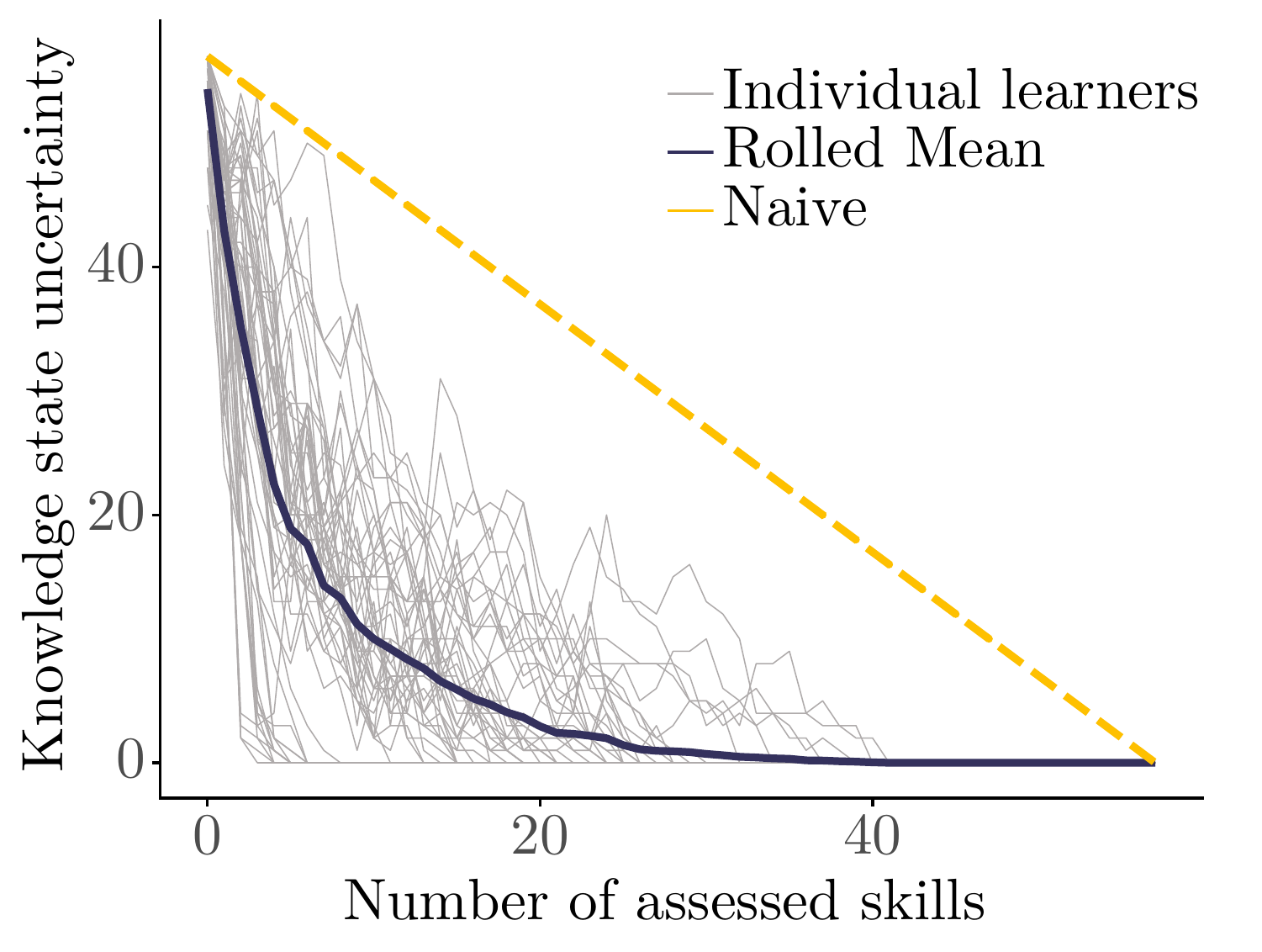}
        \subcaption{Microsoft Excel}
    \end{subfigure}  
    \caption{Knowledge state uncertainty during the assessment of three different learning paths (Microsoft Outlook, 22 Skills; The Value of Data Analysis, 38 Skills; Microsoft Excel, 57 Skills) on the edyoucated platform.
    We observe a quick and steep decrease for all three learning paths and a low number of assessed skills necessary to achieve an uncertainty of zero.}
    \label{fig:knowledge_state_uncertainty}
\end{figure*}

\subsection{Network architecture, training and parameter choice}
Every learning path requires the setup and training of a separate knowledge state network. 
In order to find an optimal network architecture, i.e., number of layers, nodes per layer, batch sizes etc., we use a standard grid search algorithm to evaluate the performance of different parameter choices. 
We then investigate the score of each architecture on an evaluation set to eventually determine the network architecture we use in the experiments (and in practice).
For all the following experiments we use k-fold cross-validation, holding out one or multiple learners depending on the number of learners available for the respective learning path for the evaluation.
To be more precise, for all the results, the architecture of the network is determined first and stays fixed throughout the experiments. 
Then, the models are trained holding out one learner each time and evaluating the assessment performance on that learner before proceeding to the next.

As a simple, first confirmation that our knowledge state network has learned the connection between skills properly we can have a look at its \textit{a-priori} probabilities, i.e., the probabilities it predicts when fed with an entirely incomplete assessment state ($\alpha(s) = 0$ for all $s \in \Scal$).
These should roughly resemble the ``true'' a-priori probabilities for mastery in our group of learners, i.e., the proportion of learners that have mastered a skill (cf. also Figure \ref{fig:user_assessment_mastery_outlook}, taking the mean over all rows). 
And indeed this is the case.
Figure \ref{fig:initial_probabilities_outlook} shows the probability output of the network vs. the a-priori mastery for the \textit{Microsoft Outlook} learning path, and we observe that for most of the skills the probabilities are similar, giving us more confidence in the network. 

We like to mention that there are, of course, numerous ways to optimize the networks and performance even further. 
In our experiments, however, we observed that the \textit{exact} network structure did not matter too much; the assessment process indeed had a similarly good quality for a broader range of architectures.
Nevertheless, an interesting approach to achieve an improvement can be, e.g., to replace the network training objective by the absolute value (instead of the square), which makes the network more likely to quickly force predicted probabilities close to zero or one. 
Due to the linear penalty instead of a squared one, this will lead to a network which takes even less assessment iterations to be certain about predictions (in the sense of Section \ref{sec:assessment_strategies}).
Note as well that, with an increasing number of training data (i.e., learners that have finished the assessment) the results are very likely to improve further, since the heterogeneity in knowledge states is covered better. 
This might as well require to increase the complexity of the network structure which, however, is not problematic due to its current simplicity.

\subsection{Full assessments}

\begin{figure*}[!t]
    \centering
    \captionsetup[subfigure]{
        labelformat=empty, 
        justification=centering
    }   
    \begin{subfigure}[c]{0.32\textwidth}
        \centering
        \includegraphics[width=\textwidth]{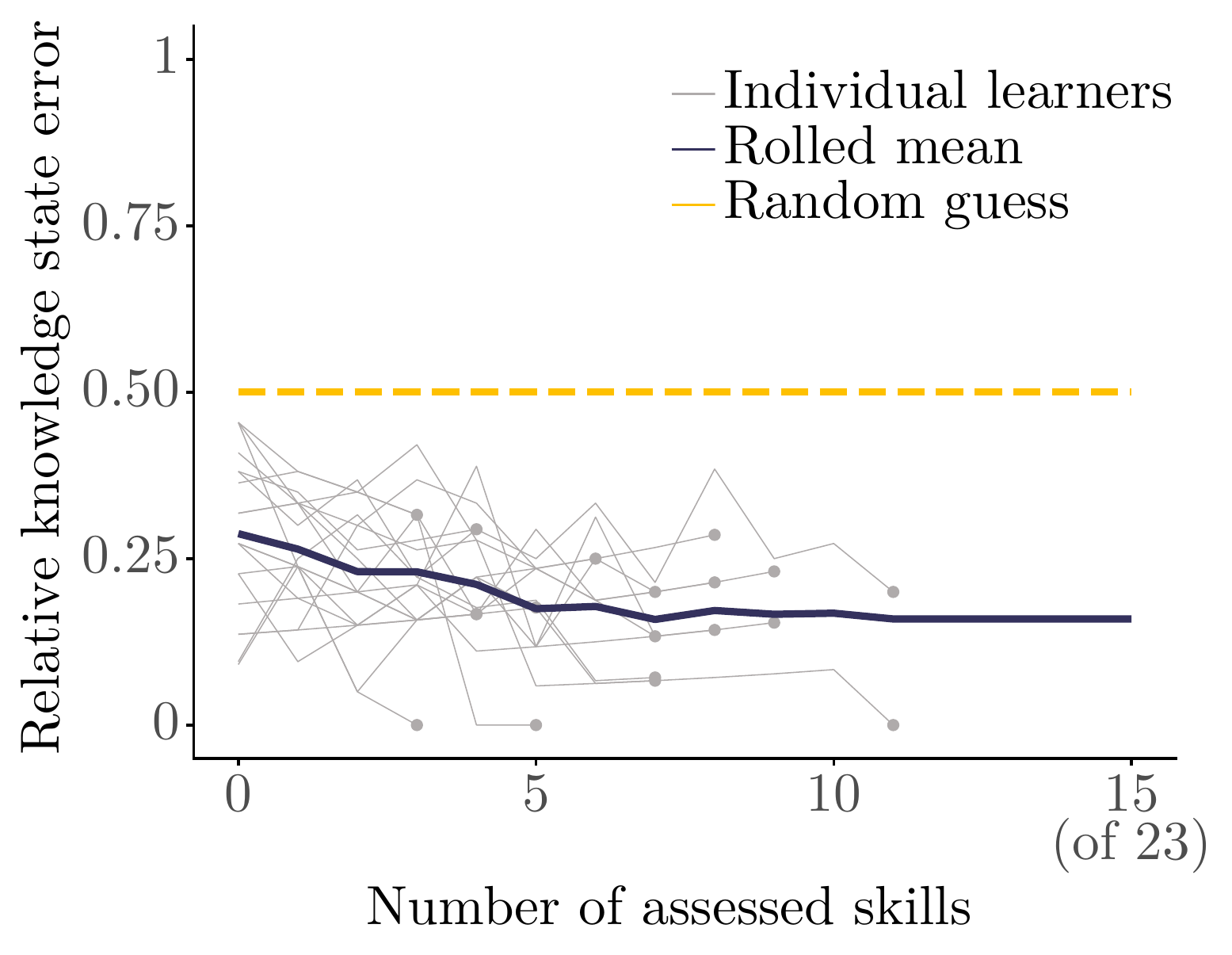}
        \subcaption{Microsoft Outlook}
    \end{subfigure}     
    \begin{subfigure}[c]{0.32\textwidth}
        \centering
        \includegraphics[width=\textwidth]{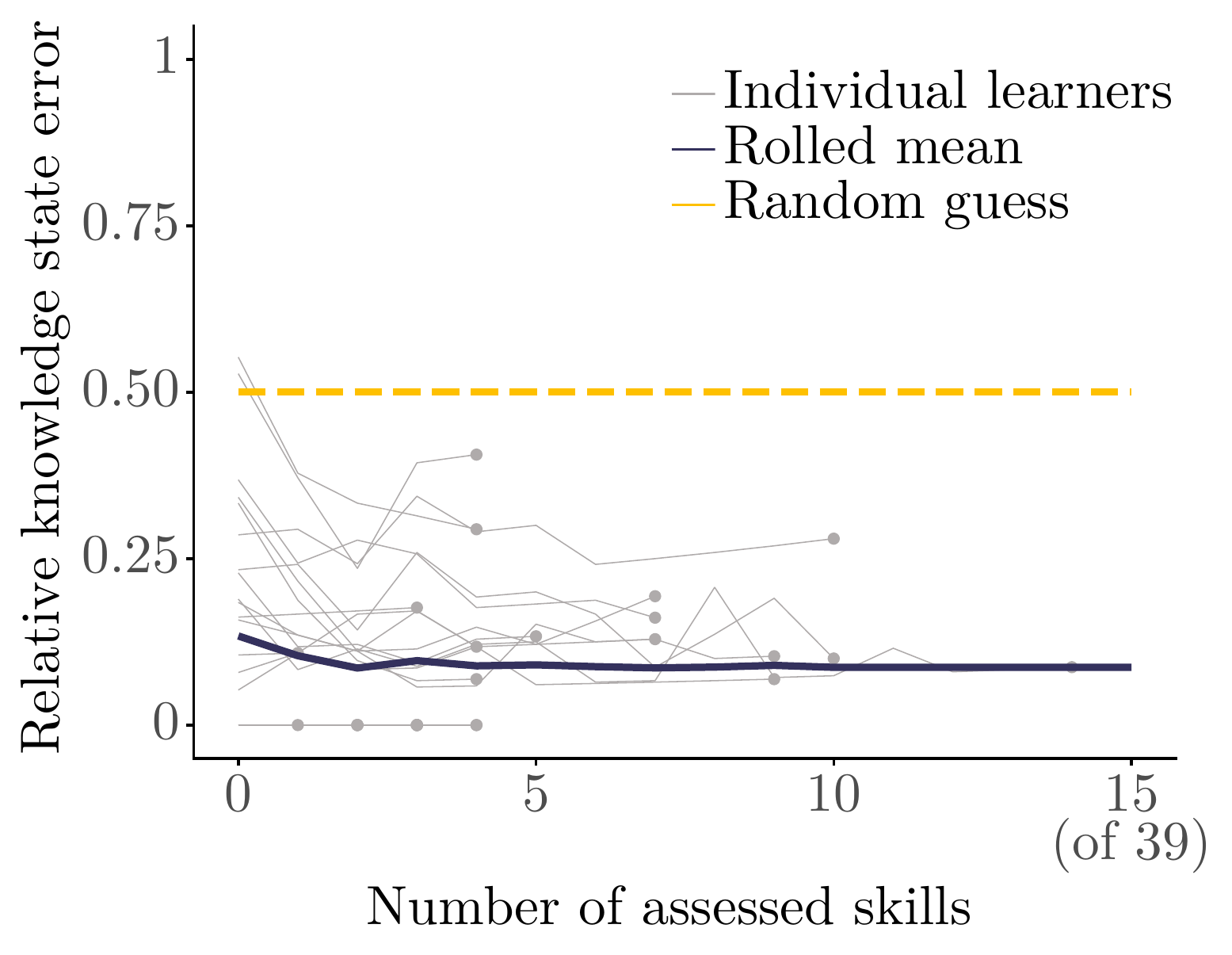}
        \subcaption{The Value of Data Analysis}
    \end{subfigure} 
    \begin{subfigure}[c]{0.32\textwidth}
        \centering
        \includegraphics[width=\textwidth]{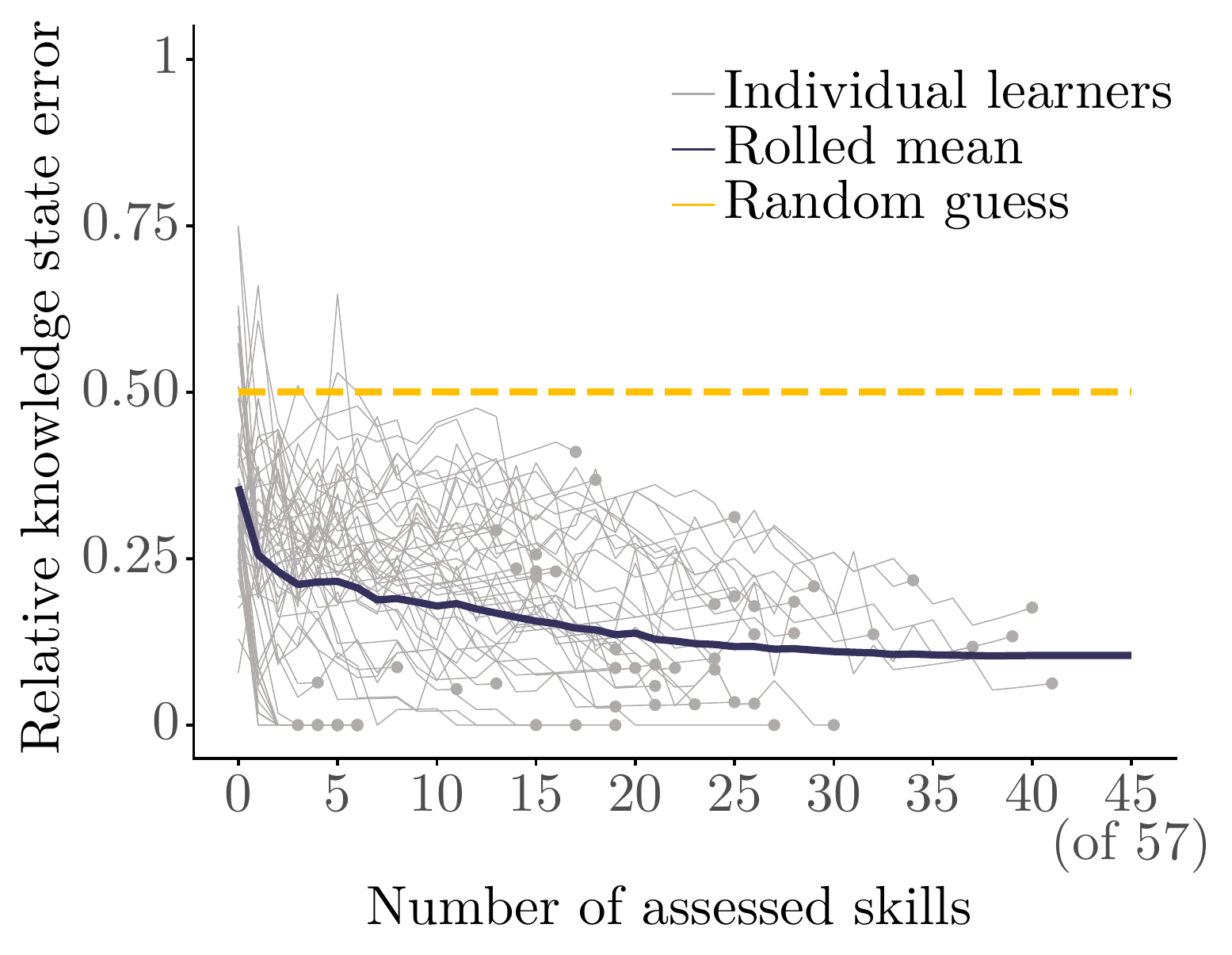}
        \subcaption{Microsoft Excel}
    \end{subfigure}  
    \caption{
        Relative knowledge state error on the \textit{unassessed} skills over the course of the assessment. 
        The rolled mean is computed by rolling the error of each individual learner forward if their assessment has already stopped.
    }
    \label{fig:relative_knowledge_state_error}
\end{figure*}
At first we want to have a look at full assessments, i.e., the mastery of all the skills of a learning path has to be figured out in a single assessment session. 
The goal is to assess as few skills as possible, while predicting the full knowledge state of the learner as precisely as possible. 
We stop the assessment once the knowledge state uncertainty error \eqref{err:knowledge_state_uncertainty} is zero, implying that the knowledge state network is ``certain'' to have figured out the true knowledge state.
As the threshold we choose $\eps = 0.1$.

\begin{table*}[!t]
    \centering
    \caption{
        Statistics for the experiments in Figures \ref{fig:knowledge_state_uncertainty} and \ref{fig:relative_knowledge_state_error}.
        Average error, standard deviation and maximum error refer to the relative knowledge state error.
    }
    \resizebox{\textwidth}{!}{
        \begin{tabular}{l c c c c c c c}
            \textbf{Learning path} & \textbf{Number of } & \textbf{Number of } & \textbf{Average} & \textbf{Maximum} & \textbf{Average} & \textbf{Standard} & \textbf{Maximum}
            \\  & \textbf{learners} & \textbf{skills} & \textbf{iterations} & \textbf{iterations} & \textbf{error} & \textbf{deviation} & \textbf{error}
            \\[2pt] \hline \hline
            \begin{filecontents*}{data/stats.csv}
Title,NumberOfLearners,AverageIterations,MaximumIterations,NumberOfSkills,AverageError,StandardDeviation,MaximumError
The Value of Data Analysis,28,5,14,38,0.09,0.11,0.41
Microsoft Outlook User,19,7,11,22,0.16,0.10,0.32
Microsoft Excel User,56,19,41,57,0.10,0.10,0.41
\end{filecontents*}
\csvreader[head to column names]{data/stats.csv}{}%
            {\\ \Title & \NumberOfLearners & \NumberOfSkills & \AverageIterations & \MaximumIterations & \AverageError & \StandardDeviation & \MaximumError}%
            \\ \hline
        \end{tabular}
    }
    \label{tbl:statistics}
\end{table*}

Note that the certainty of the algorithm does \textit{not} imply that its predictions of the knowledge state are indeed correct; it is merely the correct moment to stop the assessment since the prediction accuracy, most likely, will not increase further.
In order to evaluate the quality of the prediction for our experiments we compute the relative knowledge state error \eqref{err:relative_knowledge_state_error} on the residual (i.e., unassessed) skills after each iteration of the assessment.
For a simple algorithm which asks randomly, this error will be equal to $0.5$ on average; indeed, the knowledge about one more skill does not contribute to the remaining unassessed skills which are still equally likely to be mastered or not. 
For our algorithm we hence want to observe a relative knowledge state error which is substantially lower than $0.5$ and which tends to decrease over the course of the iterations of the assessment, implying that our predictions get better when the assessment state of the learner gets more complete.
Furthermore, it is important to keep in mind that we always evaluate the performance of the entire assessment process, i.e., the knowledge state network together with the skill picking strategy instead of the individual components since they can ever only work in unison. 
Small improvements in the prediction accuracy of the knowledge state network might have little to no effect on the assessment if it does not change the strategy of picking skills to assess.

The results of the experiments for the three different learning paths can be found in Figures \ref{fig:knowledge_state_uncertainty} and \ref{fig:relative_knowledge_state_error} as well as Table \ref{tbl:statistics}.
As we can see in Figure \ref{fig:knowledge_state_uncertainty}, the uncertainty of the network decreases rapidly after the first few assessed skills and hits zero (and hence the stopping criterion) substantially earlier than simply asking for every skill one after another (``Naive''). 
While for the shorter \textit{Microsoft Outlook} learning path we need to assess roughly a third of the skills on average (\ref{tbl:statistics}), for the longer learning paths the amount further reduces to one sixth or even less. 
The average relative knowledge state error after stopping ranges between nine percent (\textit{The Value of Data Analysis}) and sixteen percent (\textit{Microsoft Outlook}) for the remaining skills. 
To give some interpretation, e.g., for \textit{The Value of Data Analysis}, this means that after an average of five iterations we end up with an error of roughly ten percent on roughly 30 remaining skills.
This amounts to three skills that have been misassigned and 27 that have been predicted correctly. 
Of course, this is only the average performance and we are particularly interested in the learners (and outliers) where the performance is worse. 
To this end, the maximum error can as well be found in Table \ref{tbl:statistics}, which for \textit{The Value of Data Analysis} amounts to 41 percent. 
This error is reached after only four iterations (see Figure \ref{fig:user_assessment_mastery_the_value_of_data_analysis}) and hence roughly 14 skills have been predicted wrongly and 20 have been predicted correctly.  
Both a reason and a remedy are quickly found: 
As one expects, particularly bad performance in most cases correlates with learners that differ substantially from the training set and require the network to extrapolate, which is hardly possible. 
The limited amount of learners (28) and hence training data for this learning path together with the higher number of skills (38) makes it rather difficult to obtain a high performance for every learner. 
As a remedy to this we use our platform to let the learner correct the predictions of the network. 
As lined out in Section \ref{sec:assessment_strategies}, this both allows us to verify our algorithm and gather further training data and empowers the learner to learn exactly what is missing.
In the long run, the higher the number of learners of a particular learning path with diverse prior knowledge, the better the predictions will get (see Section \ref{subsec:increasing_training_data}).
Interesting to see is that, with an increasing number of skills in a learning path, the amount of assessment steps does \textit{not} increase proportionally.
The average error and standard deviation increase slightly but not unreasonably. 

Another interesting insight can be drawn from the precision and recall statistics in Table \ref{tbl:precision_and_recall}. 
Similar to the evaluation of standard classification algorithms, precision here refers to the proportion of skills predicted as mastered that has actually been mastered by a learner.
Recall indicates to the proportion of mastered skills that have been correctly identified as such by the network. 
We refer to ``negative'' precision and recall for the same measures but for skills predicted as \textit{unmastered}.
For the values presented in the table, the prediction threshold has been set to $0.5$.
The table offers insight into the inherent bias in the network. 
Depending on the learning path, learners tend to have more or less prior knowledge (mastery), and mastery naturally rather occurs in the earlier parts of the learning path. 
Hence one would expect a bias towards ``no previous knowledge'' for learning paths where the average learner has little knowledge, a bias towards ``more previous knowledge'' for learning paths where the average learner has a lot of previous knowledge, and a somewhat mixed picture when prior knowledge is distributed evenly.

And this is indeed exactly what we observe in Table \ref{tbl:precision_and_recall}.
The training for the learning path \textit{The Value of Data Analysis} features only few learners with previous knowledge, hence we recognize the bias towards ``no knowledge'': 
while negative precision and recall are high, positive recall is clearly the weakest, implying that the algorithm fails more often when trying to detect mastered skills than detecting unmastered skills. 
Note that this is exactly what we expect for the underlying data. 
For the \textit{Microsoft Excel} learning path the result is rather reversed, the bias here tends towards predicting skills as mastered. 
The learning path \textit{Microsoft Outlook} is the most balanced one, showing an equal performance in all measures. 

Of course, one can start tweaking the prediction by manipulating the prediction threshold, however, we like to take this behavior as a feature: 
Firstly, we cannot expect to perform better than our training data, i.e., if there has been no similar learner before we expect a worse performance of the algorithm.
Secondly, this inherent bias in a sense represents the learning community of the platform as well as the structure of the topic that is learned:
if the topic is difficult, the tendency will go towards no mastery, or if most learners know a topic it will tend towards the opposite. 

\begin{table}
    \centering
    \caption{
        Precision and recall statistics for the experiments in Figures \ref{fig:knowledge_state_uncertainty} and \ref{fig:relative_knowledge_state_error}.
        Negative precision and recall refer to precision and recall for the prediction of \textit{no} prior knowledge, the prediction threshold has been set to $0.5$.
    }
    \resizebox{0.6\textwidth}{!}{
        \begin{tabular}{l c c c c}
            \textbf{Learning path} & \textbf{Precision} & \textbf{Recall} & \textbf{Negative } & \textbf{Negative }
            \\  & & & \textbf{precision} & \textbf{recall}
            \\[2pt] \hline \hline
            \begin{filecontents*}{data/pr_stats.csv}
Title,Recall,Precision,NegativeRecall,NegativePrecision
The Value of Data Analysis,0.67,0.90,0.98,0.92
Microsoft Outlook User,0.84,0.81,0.81,0.82
Microsoft Excel User,0.85,0.91,0.78,0.82
\end{filecontents*}
\csvreader[head to column names]{data/pr_stats.csv}{}%
            {\\ \Title & \Precision & \Recall & \NegativePrecision & \NegativeRecall}%
            \\ \hline
        \end{tabular}
    }
    \label{tbl:precision_and_recall}
\end{table}

\subsection{Increasing the training data}
\label{subsec:increasing_training_data}

\begin{figure}[t!]
    \centering
    \includegraphics[width=0.45\textwidth]{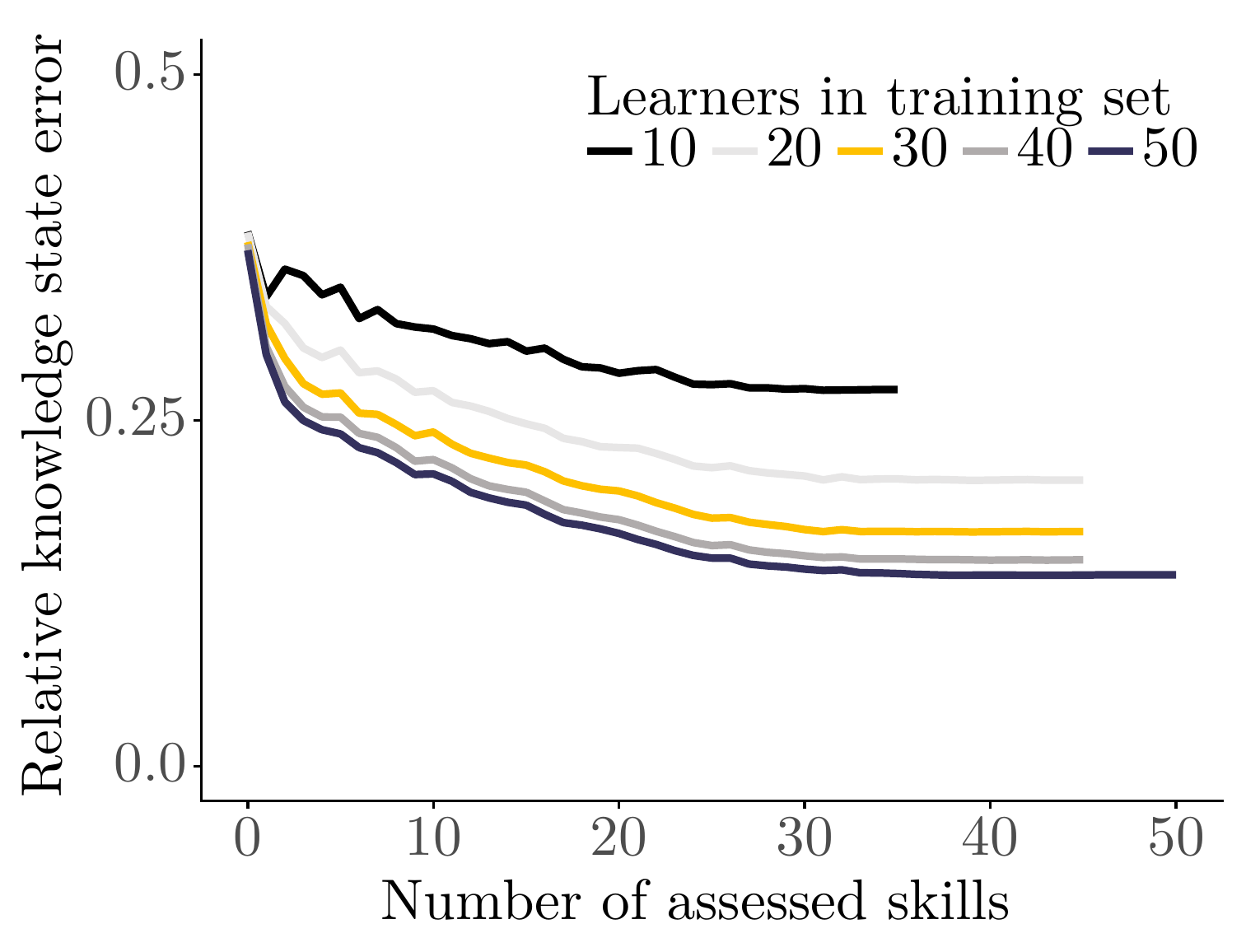}
    \caption{
        Average knowledge state error for the learning path \textit{Microsoft Excel} and different amounts $k \in \{10, 20, 30, 40, 50\}$ of learners in the training dataset.
        As expected, an increasing amount of heterogeneity (larger $k$) in the training data increases the performance of the network.
    }
    \label{fig:increasing_training_data}
\end{figure}

As a last experiment we want to demonstrate the positive effect of an increasing amount of training data for the network or, more precisely, an increasing heterogeneity in the learner base for the training. 
To this end, we use the \textit{Microsoft Excel} role as it features the highest amount of learners.
For the experiment we sample $k$ random learners from the entirety of learners, with increasing $k \in \{10, 20, 30, 40, 50\}$, and train the network on assessment states simulated from those only. 
For each $k$ we then perform the assessment for all learners as before, always excluding a learner from the training set if they are among the $k$ (in this case, only $k-1$ learners are in the training set).  
Figure \ref{fig:increasing_training_data} shows the average relative knowledge state error for each $k$ which, as expected, decreases with an increasing amount of heterogeneity in the training data (increasing $k$).
This decrease appears to be diminishing with increasing $k$ which, however, is not surprising, as additional learners do not necessarily bring more heterogeneity into the dataset. 
The more \textit{different} learners we include, the fewer errors we expect.
It is hence beneficial to include as many (different) learners in the training data in order to obtain the best possible network performance.

Of course, all the above results require further investigations that will be carried out in further studies with more learning paths and larger training sets, but we are positive the results will be similar (and even better).
It is worth mentioning that we do not include any experiments with session learning, as they depend too heavily on the ``duration'' of a session, hence introducing yet another parameter to the experiments, and the overall methodology is exactly the same.

\section{Conclusion and outlook}
In this paper, we have outlined an approach for optimized skill assessment for atomic learning, using a knowledge state network to predict the (prior) knowledge states of learners from incomplete assessments. 
We have shown that the involved neural networks are able to capture well the knowledge structure of learners in different situations which, in combination with the right assessment question strategy, significantly speeds up the assessment process for learning paths involving many (atomic) skills.
When implemented carefully, the concepts and related findings can benefit online learners in a multitude of ways. 
First and foremost, they enable us to keep up with the ever-increasing heterogeneity of the online learner base through a better personalization of learning topics and contents on an atomic level, i.e., on a very fine granularity, in order to fit the learning process best to the learners' individual needs.
This is not viable without (machine learning) models that can ``take the teacher's role'' and predict the actual knowledge states of learners from only partial information about their skill set. 
We have demonstrated that our knowledge state networks -- when trained on the right learner data -- can reliably perform this task with high accuracy. 
These networks furthermore enable us to employ automated assessment strategies which make use of the networks' predictions in order to reduce the amount of necessary assessment questions by often more than sixty percent on average.
Consequently, online learners are enabled to start their learning journey tailored to their own, individual level while avoiding an uncomfortable and lengthy assessment process; time, that can be allocated to different (learning) tasks.
As a side-effect, a short assessment process significantly reduces the ``onboarding'' process such that learners can start with the actual learning almost right away, thus maximizing the use of the generally limited time for learning and reducing the risk of drop-out directly at the beginning.
The developments of this paper hence serve as another puzzle piece for better and more effective, personalized online learning.

While our experiments and our experience on our learning platform\footnote{\url{www.edyoucated.org}} show satisfying results, there is still room for improvement in every component of the algorithm. 
Starting with the assessment state simulation for the generation of training data, one might think about deviating from the entirely uniform sampling of skills towards different probability distributions. 
One possibility, for example, is to use the a-priori probabilities of mastery (the proportion of learners that have mastered a skill) as a distribution to draw from in order to sample more relevant, incomplete assessment states for the training of the network.
A second area of improvement are the networks themselves. 
While we have pointed out in the numerical section that the results are rather robust with respect to changes in network architecture, there are always further network structures to explore for better performance.
This will be particularly interesting in combination with more learner data and even more complex learning paths which we postpone to our future studies on the topic. 
It will be interesting to see how the performance of the assessment keeps up with the increasing complexity of the assessment and knowledge states.
Furthermore, we are working on different skill picking strategies in order to further increase the performance of the assessment, but also to improve the assessment experience of the learner, for example, by producing more reasonable assessment question orders. 
Last, but not least, from a machine learning perspective, it is interesting to investigate the networks themselves and, for example, derive further relationships between atomic skills such as clusters directly from the model. 

\bibliography{references}{}
\bibliographystyle{abbrv}

\newpage
\section*{Appendix}
We have included some additional figures in this appendix. 
\begin{figure*}[h!]
    \centering
    \captionsetup[subfigure]{
        labelformat=empty, 
        justification=centering
    }   
    \begin{subfigure}[c]{0.38\textwidth}
        \centering
        \includegraphics[height=0.36\textheight]{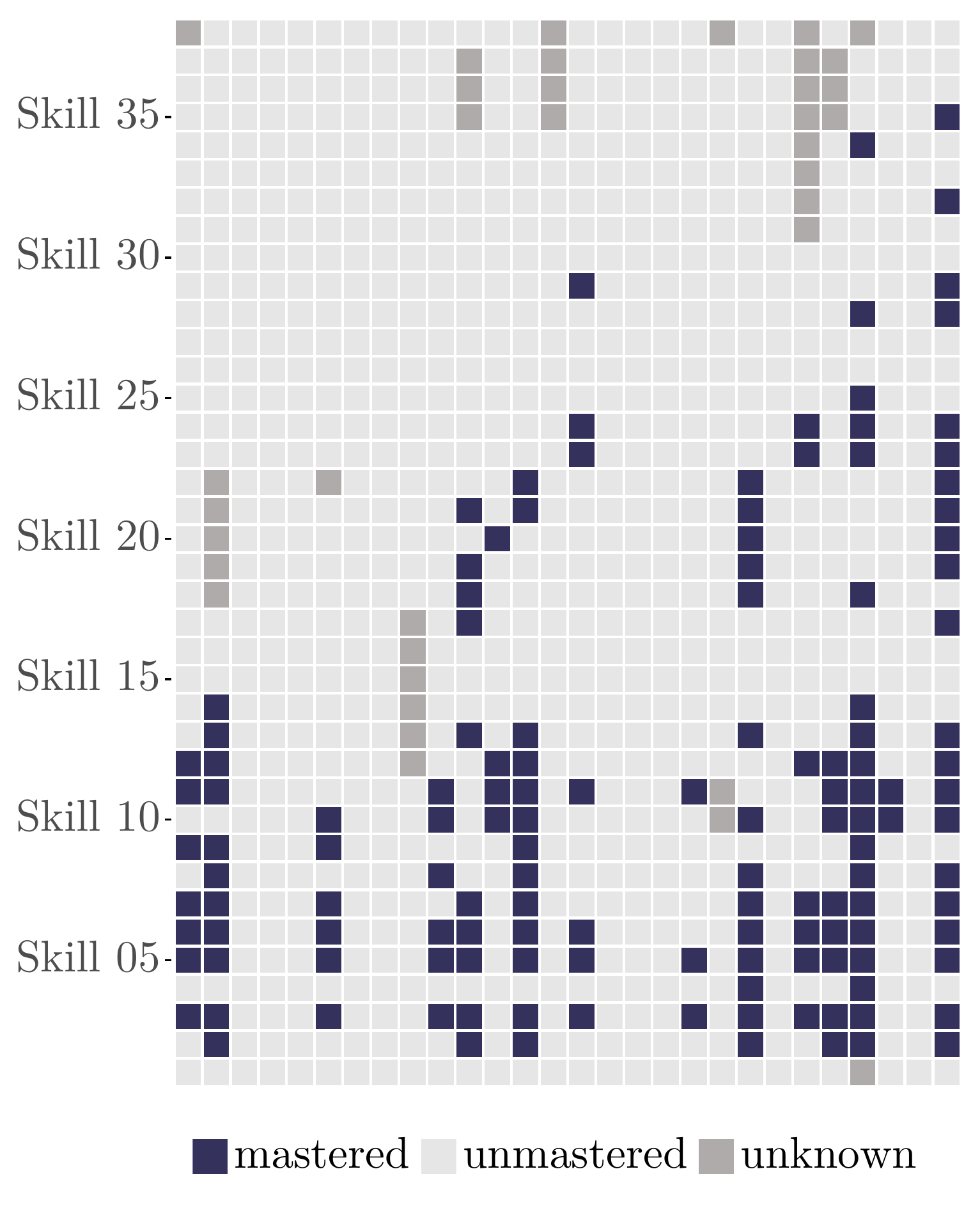}
        \subcaption{a)}
        \label{fig:user_assessment_mastery_the_value_of_data_analysis}
    \end{subfigure}%
    \hfill
    \begin{subfigure}[c]{0.6\textwidth}
        \centering
        \includegraphics[height=0.36\textheight]{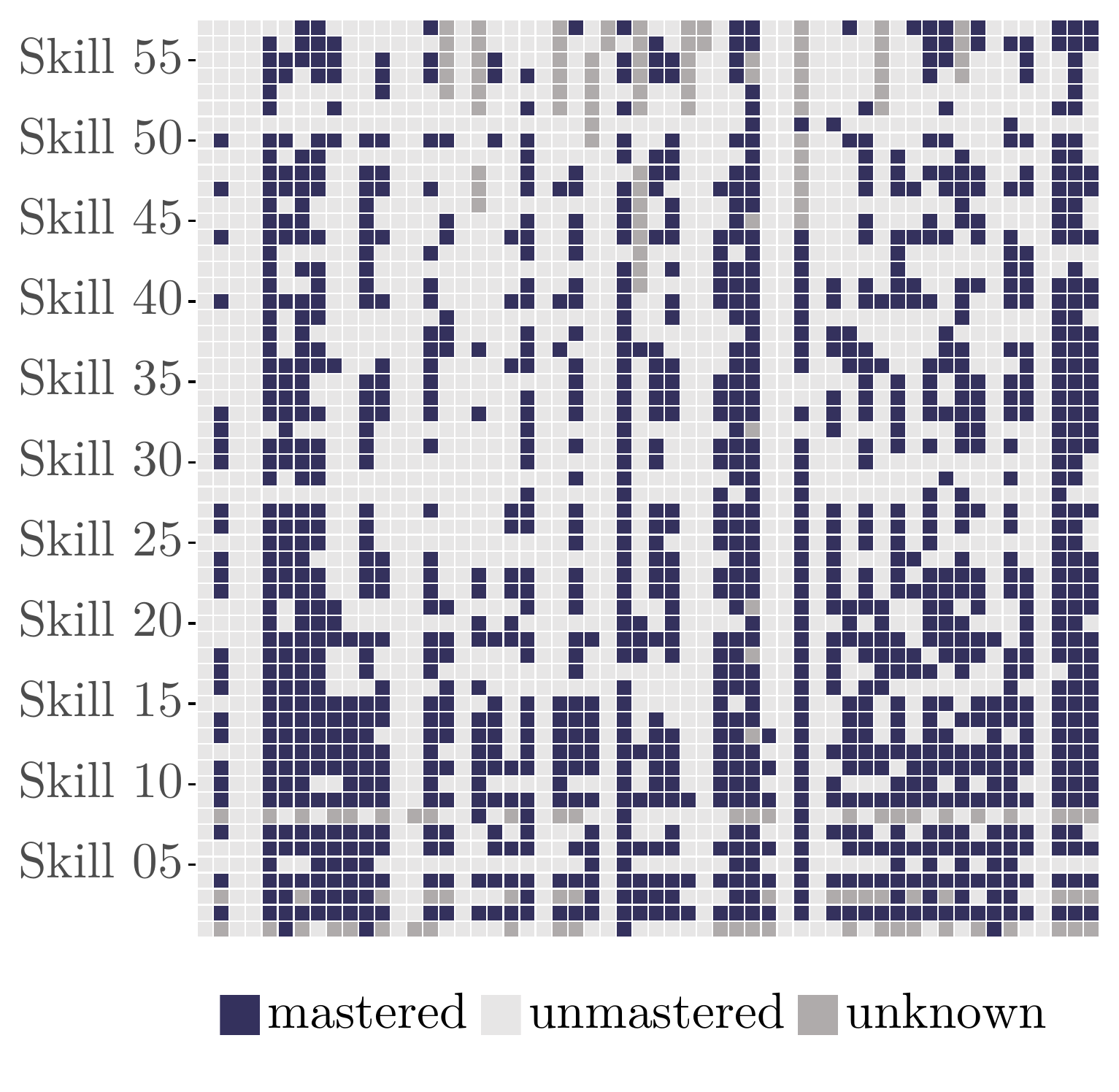}
        \subcaption{b)}
        \label{fig:user_assessment_mastery_excel}
    \end{subfigure} 
    \caption{
        Learner assessment mastery for the learning path a) \textit{The Value of Data Analysis} and b) \textit{Microsoft Excel}.
        The columns of the map indicate the different learners.
        Unknown skill mastery occurs when learners have not fully completed the assessment on the platform.
    }
    
\end{figure*}

\end{document}